\documentclass[final]{siamltex}
\usepackage{algorithm,algpseudocode}
\usepackage{latexsym, graphicx, epsfig, amsmath, amsfonts,amssymb,bm}
\usepackage{caption, subcaption}
\usepackage{epstopdf}
\usepackage{booktabs}
\usepackage{array}
\usepackage{color}
\usepackage{lineno}
\usepackage{url}
\usepackage{graphicx}
\usepackage{grffile}
\usepackage{subcaption}

\def\be{\begin{equation}}
\def\ee{\end{equation}}

\def\x{\mathbf{x}}
\def\y{\mathbf{y}}
\def\w{\mathbf{w}}
\def\f{\mathbf{f}}
\def\z{\mathbf{z}}
\def\zh{\hat{\z}}
\def\p{\mathbf{p}}
\def\q{\mathbf{q}}
\def\I{\mathbf{I}}
\def\zt{\widetilde{\z}}
\def\X{\mathbf{X}}
\def\Z{\mathbf{Z}}

\def\Rb{\mathbf{R}}
\def\Rs{\mathbb{R}}

\def\M{\mathbf{M}}
\def\A{\mathbf{A}}

\def\K{\mathbf{K}}
\def\F{\mathbf{F}}

\def\N{\mathbf{N}}
\def\Pi{\mathbf{\Phi}}
\def\PPh{\mathbf{\Phi}}
\def\PPs{\mathbf{\Psi}}

\newcommand{\argmin}{\operatornamewithlimits{argmin}}

\title{Learning Reduced Systems via Deep Neural Networks with Memory}

\author{Xiaohan Fu\footnotemark[1]\and Lo-Bin Chang\thanks{Department of Statistics,
    The Ohio State University, Columbus, OH 43210, USA. {\tt
      Emails: fu.688@osu.edu, lobinchang@stat.osu.edu.}} \and Dongbin
       Xiu\thanks{Department of Mathematics,
		The Ohio State University, Columbus, OH 43210, USA.
		{\tt Email: xiu.16@osu.edu.}
		Funding: This work was partially supported by AFOSR FA9550-18-1-0102.}
}

\begin{document}
\maketitle
\begin{abstract}
We present a general numerical approach for constructing governing equations for unknown dynamical systems when only data
on a subset of the state variables are available. The unknown equations for these observed variables
are thus a reduced system of the complete set of state variables. Reduced systems possess memory integrals, based on
the well known Mori-Zwanzig (MZ) formulism. Our numerical strategy to recover the reduced system starts by formulating a
discrete approximation of the memory integral in the MZ formulation. The resulting unknown approximate MZ equations are of finite 
dimensional, in the sense that a finite number of past history data are involved. We then present a deep neural network
structure that directly incorporates the history terms to produce memory in the network. The approach is suitable for any
practical systems with finite memory length. We then use a set of numerical examples to demonstrate the effectiveness of our
method.
\end{abstract}

% main text
\section{Introduction} \label{sec:intro}

Designing data-driven numerical methods to discover unknown physical
laws has received an increasing amount of attention lately.
Several methods were developed for dynamical systems by using traditional numerical
approximation techniques. In these approaches, the unknown governing
equations are treated as target functions, whose inputs are the state
variables and outputs are their temporal derivatives. Methods using
sparse recovery, as well as more standard polynomial approximations,
have been developed, cf. \cite{tibshirani1996regression,
brunton2016discovering, brunton2016discovering,
schaeffer2017sparse,kang2019ident, tran2017exact,
schaeffer2017extracting, rudy2017data,
schaeffer2017learning, WuXiu_JCPEQ18, WuQinXiu2019, long2017pde,raissi2017physics1,raissi2017physics2,raissi2018deep,long2018pde,sun2019neupde}. 
More recently, more research efforts are being devoted to the use of
modern machine learning techniques, particularly deep neural networks (DNNs).      
The studies include recovery of
ordinary differential equations (ODEs) \cite{raissi2018multistep,qin2018data,rudy2018deep} and partial differential equations (PDEs)
\cite{long2017pde,raissi2017physics1,raissi2017physics2,raissi2018deep,long2018pde,sun2019neupde}. 
A notable development along this line of approach is the use of flow
map for modeling the unknown dynamical equations \cite{qin2018data}. Flow map describes
the (unknown) mapping between two system states. Once it is accurately
approximated, it can serve as a model for system prediction. The major
advantage of using flow map is that it avoids requiring temporal
derivative data, which can be difficult to acquire in practice and
often subject to larger errors. In particular, residual network
(ResNet), developed in image analysis community (\cite{he2016deep}), 
is particularly suitable for equation recovery, in the sense
that it can be an exact integrator \cite{qin2018data}. This approach
has since been extended and applied to other problems
\cite{WuXiu_JCP20, QinCJX_IJUQ20, ChenXiu_JCP20}.

The aforementioned approaches are data driven and rely on
observational data of the state variables to numerically estimate the
underlying dynamical systems. 
For many practical systems, however, one does not have access to data
for all the state variables. Instead, one often only have data on a
subset of the variables, i.e., the observables. It is then natural to
seek a governing equation for the  evolution of the observed
variables.
This, however, introduces additional challenges from
mathematical point of view. Even when the underlying governing
equations for the full variable set are autonomous, the effective
governing equations for the observed variables, i.e., the reduced
system of equations, include memory terms
and become non-autonomous. This is a direct result of the well known
Mori-Zwanzig (MZ) formulation \cite{mori1965, zwanzig1973}. The memory
term in the MZ formulation represents a significant computational
challenge. Various approximation techniques have been developed to
facilitate efficient estimation of the memory effect. See, for example,
\cite{chorin2002, Bernstein2007, HaldStinis2007, Stinis2007,
  ChertockGS2008, VenturiK_2014, ZhuVenturi_2018}, and the references therein. And more recently,
data driven methods were developed to provide effective closure or
estimation for memory integral \cite{LeiBakerLi_2016,
  BrennanVenturi_2018}.

The topic of this paper is on data driven learning of unknown dynamical
systems when only data on a subset of the state variables, i.e., observables, are
available. We make a general assumption that the underlying unknown system of
complete equations are autonomous. Our goal is to construct a dynamical
model for the evolution of the observables, whose data are
available, thus discovering a reduced system. Due to the MZ
formulation, the unknown governing equations for the observables are
non-autonomous and possess memory integrals. Therefore, the
aforementoned existing
data driven methods for equation discovery are not
applicable. On the other hand, the existing approximation techniques
for the memory integral in MZ equations are not applicable either, as
the underlying complete system is unavailable.
We therefore propose a new method to directly learn the evolution
equations for the observables, with a built-in memory effect. To
accomplish this, we make a general assumption that the reduced systems
for the observables have ``decaying memory'' over longer time
horizon.  When the observables are representative of the full system
states, this usually holds true as the evolution of the observables
depends on their current states and their immediate past, and
the dependence does not usually extend to infinite past. In another
word, the initial states of the observables should have diminishing
effects on their evolution over longer time. Based on the  decaying
memory assumption, we then truncate the memory integral in the MZ
formulation up to its ``memory length'' to obtain an approximate MZ (AMZ) equation. The AMZ equation is then
discretized by using a set of time instances inside the memory
interval. The resulting discrete approximate MZ (d-AMZ) equation,
still unknown at this stage, becomes our goal of equation learning. 
We then design a deep neural network (DNN) structure that explicitly
incorporate the observable data inside the memory interval. The
proposed DNN structure is an extension of the ResNet structure used
for autonomous system learning (\cite{qin2018data}). By incorporating
data from immediate past, the new DNN can explicitly model the memory
terms in the MZ equation. We remark that our current  method has
similarity with a recent and independent work \cite{WangRH_2020},
where similar truncation and discretization of MZ formulation was
proposed. However, the work of \cite{WangRH_2020} utilizes long
short-term memory (LSTM) neural network structure to achieve memory 
effect. Our proposed DNN structure takes a much simpler form, in the
sense that it is basically a standard full connected network and does not
requires any ``gates'' as in LSTM. The new network also allows
direct conceptual and numerical connection with the ``true'' memory of
the underlying reduced systems.

This paper is organized as follows. After the problem setup in
Section \ref{sec:setup}, we present the main method in Section
\ref{sec:method}. The decaying memory assumption is first discussed in
Section \ref{sec:finite}, followed by the discrete approximate Mori-Zwanzig
formulation in Section \ref{sec:dAMZ}. The DNN structure is then
presented in Section \ref{sec:NN}, along with its data set
construction and training in Section \ref{sec:training}. 
Numerical examples are then presented in 
Section \ref{sec:examples} to
demonstrate the properties of the proposed approach. 
\section{Setup and Preliminaries} \label{sec:setup}

Let us consider a system of ordinary differential equations (ODEs),
\be \label{govern}
\frac{d\x}{dt}=\f(\x),\qquad \x(0) = \x_0,
\ee
where $\x\in\Rs^n$ are the state variables. We assume that the form of
the governing equations, which manisfests itself via $\f:\Rs^n\to\Rs^n$, is unknown.
Let $\x=(\z;\w)$, where $\z\in\Rs^d$ is the subset of
the state variables with available data, and $\w\in\Rs^{n-d}$ is the
unobserved subset of the state variables. Our goal is to construct an
effective governing equation for the observed variables $\z$.

\subsection{Data}

We assume trajectory data are available only for the observables $\z$
and not for the full set of the state variables $\x$.
Let $N_T$ be the total number of such partially observed trajectories. For
each $i$-th trajectory, we have
\be \label{Z}
\Z^{(i)} = \left\{ \z\left(t^{(i)}_k\right)\right\}, \qquad k=1,\dots, K^{(i)},
 \quad i=1,\dots, N_T,
\ee
where $\{t_k^{(i)}\}$ are discrete time instances at which the data are
available. Note that each trajectory is originated from an unknown
initial condition $\x_0^{(i)} = (\z_0^{(i)}; \w_0^{(i)})$.
For notational convenience, we shall assume a constance time step
\be \label{Delta}
\Delta \equiv t_{k+1}^{(i)} - t_{k}^{(i)}, \qquad \forall k=1,\dots,
K^{(i)}-1, \quad i=1,\dots, N_T.
\ee
We then seek to develop a numerical model for the evolution dynamics
of $\z(t)$, without data on $\w$ and the knowledge of the full model
\eqref{govern}.

\subsection{Learning of Full System}

When data on the full set of state variables $\x$ are available, the
task of recovering the full model \eqref{govern} is relatively more
straightforward. 
A number of different approaches exist.
In this paper, we adopt and modify the approach developed in
\cite{qin2018data}, which seeks to recover 
the underlying flow-map of \eqref{govern} as opposed to the
right-hand-side of \eqref{govern}.
In particular, suppose data of the full set of state
variables are available as
$$
\X^{(i)} = \left\{ \x\left(t^{(i)}_k\right)\right\}, \qquad k=1,\dots, K^{(i)},
 \quad i=1,\dots, N_T,
$$
for a total number of $N_T$ trajectories over time instances
$\{t_k^{(i)}\}$ with a constant step size, $\Delta=t_{k+1}^{(i)}-t_{k}^{(i)}$ for all $i$ and $k$.  One can then
re-group the data into pairs of two
adjacent time instances, for each $i=1,\dots, N_T$,
$$
\left\{\x\left(t_k^{(i)}\right), \x\left(t_{k+1}^{(i)}\right)\right\}, \qquad k=1,\dots, K^{(i)}-1,
 \quad i=1,\dots, N_T.
 $$
 Note that for autonomous system \eqref{govern}, time $t$ can be
 arbitrarily shifted and only the relative time difference is
 relevant. One can then define the data set as
 \be \label{dataset0}
 \left\{\x_j(0), \x_j(\Delta)\right\}, \qquad j=1,\dots, J,
\ee
 where the total number of data pairs $J=(K^{(1)}-1)+\cdots+(K^{(N_T)}-1)$.

On the other hand, the autonomous full system \eqref{govern} defines a
flow map $\PPh:\Rs^{n}\to\Rs^{n}$, such that
$
\x(s_1) = \PPh_{s_1-s_0}(\x(s_0)). 
$
We then have
\begin{equation} \label{flowmap}
  \begin{split}
\x(\Delta) &=\x(0)+\int_{0}^{\Delta} \f(\x(s)) ds\\ &=\x(0)+\int_{0}^{\Delta}
\f(\PPh_s(\x(0))) ds \\
&=\left[\mathbf{I}_n + \PPs(\cdot)\right](\x(0)),
\end{split}
\end{equation}
%Therefore,
%\begin{equation}
 % \x(\Delta)=\left[\mathbf{I}_d + \PPs(\cdot)\right](\x(0)),
%\end{equation}
where $\mathbf{I}_n$ is the identity matrix of size $n\times n$, and for any $\x\in\Rs^n$,
$$
\PPs(\x) = \int_{0}^{\Delta}
\f(\PPh_s(\x)) ds.
$$
Based on \eqref{flowmap}, it was then proposed in \cite{QinWuXiu2019} to use
residue network (ResNet) (\cite{he2016deep}) to recover the system. The
ResNet has a structure of
    \begin{equation}\label{resnet}
      \y^{out} =  \left[\mathbf{I}_n +
        \N\right]\left(\y^{in}\right),
    \end{equation}
    where $\N:\Rs^{n}\to\Rs^n$ is the operator corresponding to
    a fully connected deep neural network. Upon using the data set
    \eqref{dataset0}, the ResNet \eqref{resnet} can be trained to
    approximate the dynamics \eqref{flowmap}, with the
    deep network operator $\N\approx \PPs$.
    
\subsection{Mori-Zwanzig Formulation for Reduced System}

The approach in the previous section, along with most other existing
equation recovery methods, does not apply to the
problem considered in this paper. The reason is because here we seek
to develop/discover the dynamic equations for only the
observables $\z$, which belong to a subset of the full set variables $\x$. Even
though the full system \eqref{govern} is autonomous, a crucial property
required in most of the existing equation recovery methods,
the evolution equations for the subset variables $\z$ become
non-autonomous. This is well understood from Mori-Zwanzig (MZ)
formulation (\cite{mori1965},\cite{zwanzig1973}).
%By following the exposition in \cite{chorin2002},
The evolution of the
reduced set of variables $\z$ follows generalized Langevin equation in
the following form,
\be \label{eq:MZ}
\frac{d}{dt}\z (t)=\Rb(\z(t)) + \int_0^t\K(\z(t-s),s)ds + \F(t, \x_0).
\ee
The first term $\Rb$ depends only on the reduced variables $\z$ at the current
time and is Markovian. The second term, known as the memory, depends
on the reduced variables $\z$ at all time, from the intial time  $s=0$ to the current
time $s=t$. Its integrand involves $\K$, commonly known as the memory
kernel. The last term is called orthogonal dynamics, which depends
on the unknown initial condition of the entire variable $\x(0)$ and is
treated as noise. Note that this formulation is an exact
representation of the dynamics of the observed variables $\z$.
The presence of the memory term makes the system non-autonomous and
induces computational challenges.
We remark again that, even though various techniques exist to estimate
the memory intergal (cf. \cite{chorin2002, Bernstein2007, HaldStinis2007, Stinis2007,
  ChertockGS2008, VenturiK_2014, ZhuVenturi_2018}), they rely on the
knownledge of the full system \eqref{govern} and are thus not
applicable in the setting of this paper, where the full system is
unknown.

%$\K$,$\F_i$ are functions, $i=1,...,d$, and the exact forms of the functions can be found in \cite{chorin2002}. The three terms on the right hand side have conventional interpretations. The first term is a Markovian term, which only depends on the value of the retained variables at current time. The second term involves the memory of the reduced system. The vector-valued function $\K$ is called memory kernel. The last term requires the knowledge of initial value of the full system, but it is usually viewed as noise. 

%\eqref{eq:MZ} is an identity, but it requires the memory of the reduced system from time 0, which is computationally expensive. So approximations and truncation of the memory are needed. To do this, we make the following assumption:

\section{Main Method} \label{sec:method}

In this section, we discuss the detail of our proposed numerical
method. We first present finite memory approximation to the exact
Mori-Zwanzig formulation \eqref{eq:MZ}. We then discuss discrete
approximation to the finite memory MZ formulation. Our neural network
model is then constructed to approximate the unknown discrete approximate MZ formulation.

\subsection{Finite Memory Approximation} \label{sec:finite}

We make a basic assumption in the memory term of the Mori-Zwanzig
formulation \eqref{eq:MZ}. That is, we assume the memory kernel $\K$
in the memory term decays over time, i.e., for sufficiently large $t>0$,
\be
\left|\K(\z(t-s),s)\right| \searrow, \qquad \text{ as } s\nearrow. 
\ee
More specifically, this is defined as follows.
\begin{definition}[Decaying memory] \label{assumption}
A reduced system \eqref{eq:MZ} is said to have uniformly decaying memory if, for
any $\epsilon>0$, there exists a nonnegative constant $T_M$, which
depends on $\epsilon$ but not on $t$, such that for all $t\ge T_M$,
\be \label{memory}
\left|\int_0^t\K(\z(t-s),s)ds - \int_0^{T_M}\K(\z(t-s),s)ds\right|\le\epsilon.
\ee
\end{definition}
For any small $\epsilon$, the decaying memory
assumption implies that the memory term in \eqref{eq:MZ} depends only
on the reduced variables $\z$ from its current state at time $s=t$ to
its recent past up to $s=t-T_M$. The memory effect from the ``earlier
time'' $s\in [0,
t-T_M)$ is negligible, up to the choice of $\epsilon$.
% The memory kernel decays over time, that is, for $s\le t$,
We remark that this assumption holds true for many practical physical
systems, whose states are usually dependent upon their immediate past
and do not extend indefinitely backward in time. In another word,
the initial conditions have diminishing influence on
the system states as time evolves. The constant $T_M$ is independent
of $t$ and is called memory length. Its value is obviously problem
dependent. Note that at the early stage of the system evolution when
time $t$ is small, i.e., $t<T_M$, the condition \eqref{memory} is
trivially satisfied by letting $T_M\leftarrow \min(t, T_M)$. (Or, by setting $\z(t) =0$, for $t<0$.)
Therefore, throughout the rest of the paper we shall use $T_M$ without
explicitly stating out this trivial case.

%Assumption 1 says that the memory that is far away from the current time has little effect on the current system, so we can truncate the memory term in \eqref{eq:MZ}. After truncation, the integral only involves $\z(t-s)$ for $0\le s\le m_l$. So computing this term only requires memory of the reduced system from time $t-m_l$ to $t$. This assumption is valid and reasonable as it is usually the case in physics. \change{reason}

Upon adopting the  decaying memory assumption \eqref{memory}, we
define the following approximate Mori-Zwanzig (AMZ) dynamical system
\be \label{AMZ}
\frac{d}{dt}{\zh} (t) = \Rb({\zh}(t)) + \int_0^{T_M}\K({\zh}(t-s),s)ds,
\ee
where $T_M$ is the memory length for a chosen small error threshold
$\epsilon$ from \eqref{memory}, and the noise term $\F$ in \eqref{eq:MZ} is embedded in
the approximation.

\subsection{Discrete Finite-Memory Approximation} \label{sec:dAMZ}

We now consider discrete representation of the AMZ system
\eqref{AMZ}. Let $\zh_n=\zh(t_n)$ be the solution at time $t_n =
n\Delta$ over a constant time step $\Delta$. Let  $T_M>0$ be the
memory length in the AMZ equation \eqref{AMZ}. Let $n_M\geq 1$ be the number
of time levels inside the memory range of time $t_n$, i.e.,
for $n_M=\lfloor T_M/\Delta\rfloor$ such that
$$
\left\{t_{n-{n_M}}, \dots, t_{n-1}\right\} \subset [t_n-T_M, t_n),
\quad
t_{n-{n_M}-1} \notin [t_n-T_M, t_n).
$$
Note that we do not count the ``current'' time level $t_n$ as part of
the memory.
We then assume that there exists a finite dimensional function $\M$
such that, for any $t_n$,
\be \label{delta}
\left|\M(\zh_{n-n_M}, \dots, \zh_{n-1}, \zh_n) -
  \int_0^{T_M}\K({\zh}(t_n-s),s)ds\right|\leq \eta(t_n; T_M, n_M),
\ee
where $\eta\geq 0$ is the error. That is, we have assumed that the
finite memory integral in AMZ \eqref{AMZ} can be approximated by the
$(n_M+1)$ dimensional function $\M$. Note that this merely
assumes a finite integral can be approximated by using the values of
its integrand at a set of discrete locations inside the integraton
domain. This is a very mild assumption used in the theory of numerical
integration, which is a classic numerical analysis topic. For example,
one can always choose $\M$ to be a certain numerical
quadrature rule using the nodes $\{\zh_{n-n_M}, \dots, \zh_{n-1},
\zh_n \}$ and with
established error behavior. Our approach in this 
paper, however, does not use any pre-selected
approximation rule for the memory integral. Instead, we shall leave
$\M$ unspecified and treat it as unknown.

With the discrete approximation to the memory integral in place, we
now define a discrete approximate Mori-Zwanzig (d-AMZ) equation,
\be \label{DAMZ}
\left.\frac{d}{dt}{\zt} (t)\right|_{t=t_n} = \left.\Rb({\zt}(t))\right|_{t=t_n} + \M(\zt_{n-n_M}, \dots, \zt_{n-1}, \zt_n),
\ee
where $\M$ is defined in \eqref{delta}. Obviously,
this is an approximation to the AMZ \eqref{AMZ} at time
level $t=t_n$, where the approximation error stems from the use of
\eqref{delta}. Note that both $\Rb$ and $\M$ on the right-hand-side
are unknown at this stage.

\subsection{Neural Network Structure} \label{sec:NN}

The d-AMZ formulation \eqref{DAMZ} serves
as our foundation for learning the dynamics of the reduced variables
$\z$. It indicates that the evolution of $\z$ at any time $t_n$
depends on not only its current state $\z_n$ but also a finite $n_M$
number of its past history states \{$\z_{n-1}, \dots, \z_{n-n_M}\}$, where
the number $n_M$ depends on the memory length $T_M$ and the time step
size $\Delta$. Based on this, we propose to build a deep neural
network structure to create a mapping from $\z_n, \z_{n-1}, \dots,
\z_{n-n_M}$ to $\z_{n+1}$ and utilize the observational data on $\z$ to
train the network.

\begin{figure}
	\centering
	\includegraphics[scale=0.4]{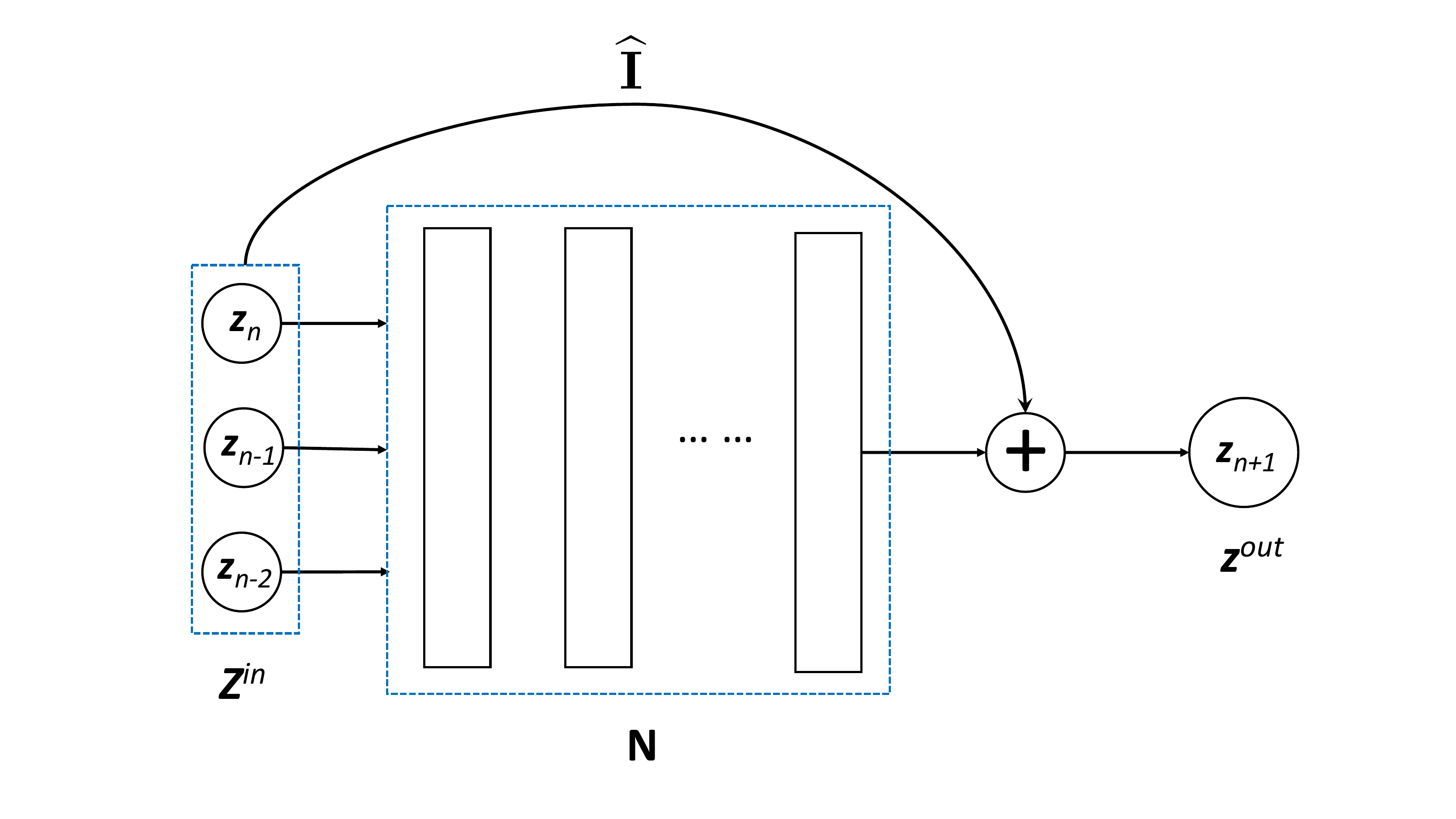}
	\caption{Illustration of the proposed neural network with
          $n_M=2$ memory steps.}
	\label{fig:Net}
      \end{figure}

Let us define
\be \label{D}
D= d\times (n_M+1),
\ee
\be \label{ZZ}
\Z =\left(\z_n^\top, \z_{n-1}^\top, \dots, \z_{n-n_M}^\top\right)^\top \in \Rs^{D}.
\ee
and a $(d\times D)$ matrix
$$
\widehat{\mathbf{I}} = \left[\mathbf{I}_d, \mathbf{0}, \dots,
  \mathbf{0}\right],
$$
%\begin{equation}
  %\widehat{\mathbf{I}} = \left[
    %  \begin{array}{cc}
      %  \mathbf{I}_d & \mathbf{0} \\
        %\mathbf{0} & \mathbf{0}
     % \end{array}
     % \right],
    %\end{equation}
    where a $(d\times d)$ size identity matrix $\mathbf{I}_d$ is
    concatenated by $n_M$ zero matrices of size $(d\times d)$. 
    Let
    \be
    \N(\cdot; \Theta):\Rs^D\to \Rs^d
    \ee
    be the operator of a fully connected feedforward neural network
    with parameter set $\Theta$. We then define a deep neural network
    in the following manner
    \begin{equation} \label{Net}
      \z^{out} =  \left[\widehat{\mathbf{I}} +
        {\N}\right]\left(\Z^{in}\right).
    \end{equation}
    A illustration of this network with $n_M=2$ memory terms is shown
    in Fig.~\ref{fig:Net}.
    %%%%%%%%%%%%%%%%%%%
      %%%%%%%%%%%%%%
It is straightforward to see that this network creates a mapping
\be \label{Net_scheme}
\z_{n+1} = \z_{n} + \N(\z_n, \z_{n-1}, \dots, \z_{n-n_M}; \Theta), \qquad n\geq n_M.
\ee

\subsection{Data, Network Training and Prediction} \label{sec:training}

To train the network \eqref{Net}, we re-organize the data set
\eqref{Z}. Let $T_M$ be the memory length. With time step $\Delta$
chosen, we have $n_M=\lfloor T_M/\Delta \rfloor$.

For the trajectory data \eqref{Z} on $\z$, let us
consider each of the $i$-th trajectory data, where $i=1,\dots, N_T$.  We
assume $K^{(i)}\geq n_M+2$. That is, each trajectory needs to
contain at least $(n_M+2)$ data entries. (Otherwise this
trajectory is discarded.) We then select a sequence of $(n_M+2)$ data
entries of consecutive time instances along this trajectory and group
them into two vectors, with the first one as the concatenation of the first $(n_M+1)$
entries in the form of \eqref{ZZ} and the second one as the last entry, in the following form,
\be \label{group}
\left\{ \Z^{(i)}_j, \z^{(i)}_j\right\}, \qquad j=1,\dots, J^{(i)},
 \quad
 \ee
 where $J^{(i)}$ is the number of such selected sequences of length $(n_M+2)$, and
\be
\begin{split}
  \Z^{(i)}_j &= \left[\z\left(t^{(i)}_{k_j}\right)^\top, \dots, \z\left(t^{(i)}_{k_j+n_M}\right)^\top,
  \z\left(t^{(i)}_{k_j+n_M+1}\right)^\top\right]^\top \in\Rs^D,  \\
\z^{(i)}_j &= \z\left(t^{(i)}_{k_j+n_M+2}\right) \in \Rs^d.
\end{split}
\ee
Here $k_j$ is the ``starting'' position of this sequence in the $i$-th
trajectory. Obviously,
when the total number of data entries along the trajectory is exactly $(n_M+2)$, the
starting position has to be $k_j=1$ and the number of such
groupings is $J^{(i)}=1$.

When the number of data entries is more than $(n_M+2)$, one may choose
$J^{(i)}>1$ number of such groups. We here discuss two straightforward
selections.
\begin{itemize}
  \item Deterministic selection. This is done by selecting the
    starting position sequentially from $k_j=1$ to $k_j= K^{(i)} -
    n_M-1$, and then for each selected starting position take a
    sequence of $(n_M+2)$ to form the group \eqref{group}. This
    results in $J^{(i)}= K^{(i)} - n_M-1$ number of groups.
  \item Random selection. Choose $J^{(i)}< K^{(i)} - n_M-1$ as the
    number of selected groups. Randomly selected $J^{(i)}$ starting
    position from the index set $\{1,\cdots, K^{(i)} - n_M-1\}$. And
    for each selected starting position, form a group in the form of \eqref{group}.
  \end{itemize}
In our numerical studies, we have found the random selection to be
more effective than the deterministic selection.

The aforementioned group selection  procedure is then repeated for
each $i=1,\dots, N_T$, trajectory. We then form the training data set
by collecting all the groups from \eqref{group} 
\be \label{data_set}
\mathcal{Z} = \left\{ \Z_j, \z_j\right\}, \qquad j=1,\dots, J,
 \quad
 \ee
 where $J= J^{(1)}+\cdots + J^{(N_T)}$ is the total number of data
 groupings. This is the training data set, where we have re-labeled
 the entries using a single
 index $j$ for computational convenience.

 By using the data set \eqref{data_set} and our network structure
 \eqref{Net}, we then train the neural network by finding its
    parameter set $\Theta^*$ that minimizes the mean-squared loss, i.e.,
    \begin{equation} \label{eq:loss}
      \Theta^* =\argmin_\Theta\frac{1}{J}\sum_{j=1}^J\left\|
        \z^{out}_j(\Theta)-\z_j\right\|^2,
    \end{equation}
    where
    $$
     \z^{out}_j(\Theta) =  \left[\widehat{\mathbf{I}} +
       {\N}(\cdot, \Theta)\right]\left(\Z_j\right), \qquad j=1,\dots, J,
     $$
     is the network output via \eqref{Net} for input $\Z_j$.
    Upon finding the optimal network parameter $\Theta^*$, we obtain
    a trained network model
        \begin{equation} \label{TheNet}
      \z^{out} =  \left[\widehat{\mathbf{I}} +
        {\N}(\cdot, \Theta^*)\right]\left(\Z^{in}\right).
    \end{equation}
 This in turn defines a predictive model for the unknown dynamical
 system for the observed variables $\z$,
 \be \label{Model}
\left\{
\begin{split}
&\z_{n+1} = \z_{n} + \N(\z_n, \z_{n-1}, \dots, \z_{n-n_M}; \Theta^*), \qquad n\geq n_M, \\
& \z_n = \z(t_n), \qquad n=0,\dots, n_M-1.
\end{split}
\right.
\ee
 With $n_M$ initial data on $\z$, one can iteratively apply the network model
 to predict the evolution of $\z$ at later time.

It is worthwhile to discuss the difference between the trained network
model \eqref{Model} and Euler forward approximation of the d-AMZ equation \eqref{DAMZ}.
If the operators $\Rb$ and $\M$ in \eqref{DAMZ} are known, its Euler
forward approximation takes the following form,
\be \label{Euler}
\left\{
\begin{split}
&\zt_{n+1} = \zt_n + \Delta\left[\Rb(\zt_n) + \M(\zt_{n},
  \zt_{n-1}, \dots, \zt_{n-n_M})\right], \qquad n\geq n_M, \\
& \zt_n = \z(t_n), \qquad n=0,\dots, n_M.
\end{split}
\right.
\ee
This obviously induces temporal discretization error. In this case
of Euler forward the error is $O(\Delta)$.

Although our model \eqref{Model} and the Euler approximation
\eqref{Euler} resemble each other, we emphasize that they 
are fundamentally different. Our neural network model \eqref{Model} is 
a direct nonlinear approximation to the time average of the
right-hand-side of \eqref{DAMZ}, whereas the Euler method
\eqref{Euler} is a pre-selected piecewise constant approximation to
the time average. Therefore, the Euler method requires the knowledge
of the operators $\Rb$ and $\M$, which is not available in our setting,
and has $O(\Delta)$ temporal
error. Our neural network model \eqref{Model}, on the other hand, does
not contain this temporal error and uses data to directly approximate
the operators in \eqref{DAMZ}. For autonomous dynamical systems, it
was shown that the neural network model is exact in temporal
integration (\cite{QinWuXiu2019}), with the only source of errors
being the training error \eqref{eq:loss}. Error analysis for the
reduced system model \eqref{Model} in this paper is considerably more complicated, and
will be pursued in separate studies.
\section{Numerical Examples} \label{sec:examples}

In this section, we present numerical examples to examine the
performance of the proposed learning method.
Our examples include two linear systems, where the exact Mori-Zwanzig
formulation for the reduced system is available, and two nonlinear
systems, one of which is chaotic. Since in all examples the true
models are available, we are able to compute their solutions with high
resolution numerical solver. This creates reference solutions, with
which we compare the predictive results by our neural network models. 
For the chaotic system, an analytically defined reduced model
is also available, and its results are used to compare against those
of our trained reduced network model.

The training data for the reduced variables are synthetic and
generated by solving the true systems with high resolution. 
In each example, we
first choose a range of interest for the full variables $\x=(\z;\w) \in
D_{\x}$. This will be the range in which we seek an accurate model for
the observed variable $\z \in D_{\z}\subset D_{\x}$. We then randomly
generate $N_T$ number of initial conditions using the uniform distribution
on $D_\x$. For each $i=1,\dots, N_T$, we solve the underlying true
system of equations with high resolution and march forward in time with
time step $\Delta$. In all examples, we set $\Delta = 0.02$.
Each trajectory is marched forward in time for $K^{(i)} = K =
\const$ steps. We then only keep the trajectory data for the observed
variables $\z$ to create our raw data set \eqref{Z}. For benchmarking
purpose, we did not add additional noises to the data. This allows us
to examine more closely the properties of the method.

The memory length $T_M$ \eqref{memory} is problem dependent. In each
example, we progressively increase the memory length to achieve
converged results. The number of memory steps is then determined as
$n_M = \lfloor T_M/\Delta \rfloor$. We then randomly select $J^{(i)} $ number
of sequences of length $(n_M+2)$ data entries from each of the $i$-th
trajectory data, where $i=1,\dots, N_T$, as described in
Section \ref{sec:training}, to form the training data set
\eqref{data_set}. We fix $J^{(i)} = J_0$ to be a constant for all
trajectories. The total number of data entries in the training data
set \eqref{data_set} is then $J= J_0 \times N_T$. In all the tests, we
purposefully keep the number of data $J$ to be roughly $5\sim 10$
times of the
number of parameters in the neural network structure. This is to avoid
any potential training accuracy loss due to lack of data and/or
overfitting, thus allowing us to focus on the properties of the
numerical methods. In practical computations when data are limited,
proper care needs to be taken during network trianing. This is well
recognized and well studied topic outside the scope of
this paper.

\subsection{Example 1: Small Linear System}

We first consider a simple linear system
\begin{equation}
\begin{cases}
\dot x_1 = x_1 - 4x_2,\\
\dot x_2 = 4 x_1 - \alpha x_2,
\end{cases}       
\end{equation}
where $\alpha$ is a parameter controlling the decay rate of the
solution.
We set $z=x_1$ to be the observed
variable and $w=x_2$ to be the unobserved variable.
Note that this is a simplified case of the well documented case
\begin{equation} \label{general}
\begin{cases}
\dot \z = \A_{11} \z + \A_{12} \w,\\
\dot \w = \A_{21} \z + \A_{22} \w,
\end{cases}       
\end{equation}
where the matrices $\A$'s are of proper sizes. The exact Mori-Zwanzig
dynamics for the observed variable $\z$ is known as
\begin{equation}\label{linearMZ}
\frac{d\z}{dt} = \A_{11} \z + \A_{12}\int_0^t e^{\A_{22}(t-s)} \A_{21}
\z(s)ds + \A_{21} e^{\A_{22} t} \w(0).
\end{equation}

In our example, we set the domain of interest to be $D_{\x}=
[-2,2]^2$. For data generation, we set all trajectory length to be
$(n_M+2)$. Hence, each trajectory contributes to one data entry ($J_0=1)$ in the
training data set \eqref{data_set}.  Based on the solution behavior, two cases are presented here: (1) Fast
decay case for $\alpha=2$; and (2) Slow decay case for $\alpha=1.1$.

\begin{figure}
	\centering
	\begin{subfigure}[b]{0.5\textwidth}
		\includegraphics[width=\textwidth]{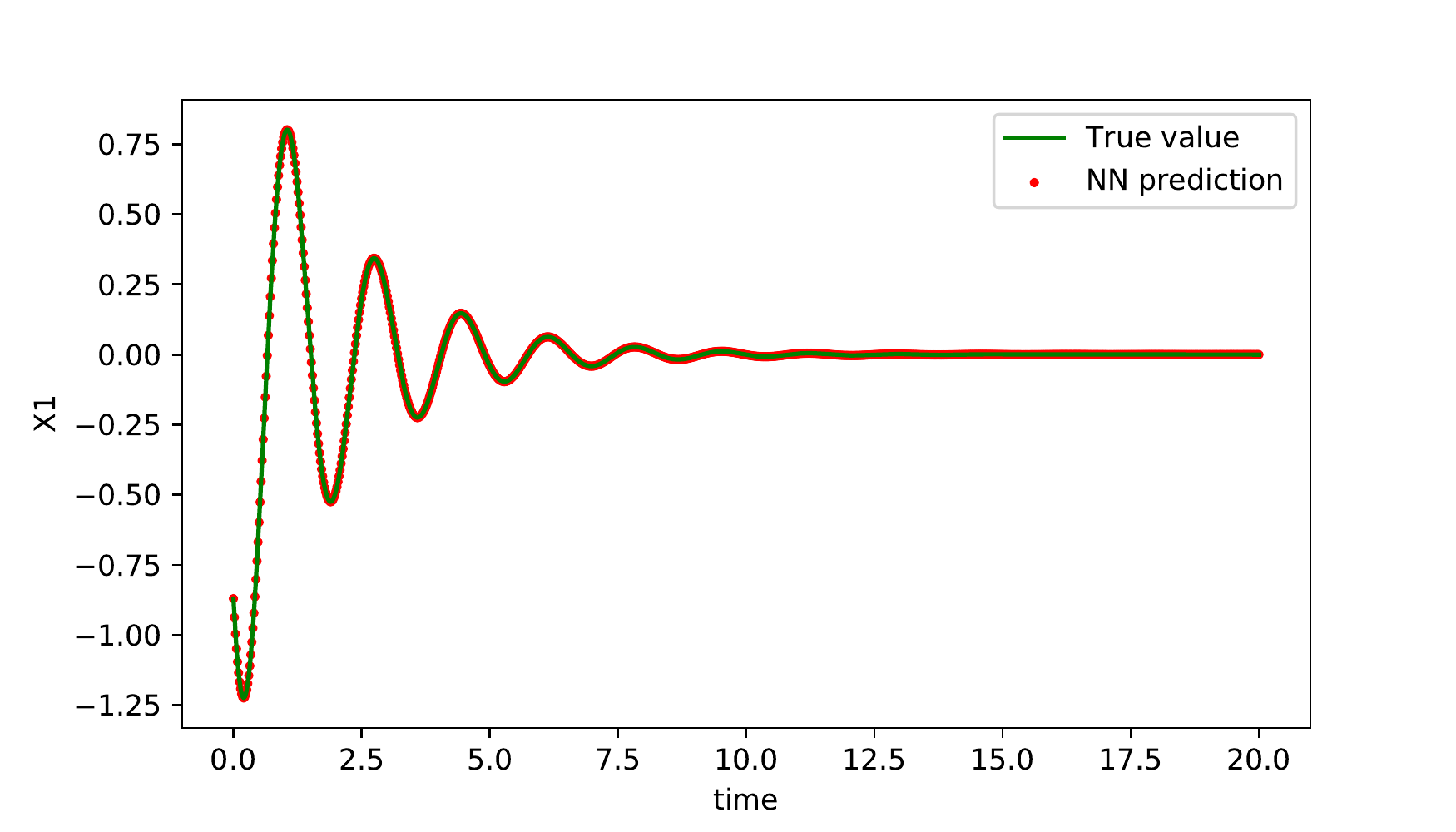}
		\caption{Initial value = $(-0.87,0.65) $.}
		\label{fig:e11_1}
	\end{subfigure}%
	%~ %add desired spacing between images, e. g. ~, \quad, \qquad, \hfill etc. 
	\begin{subfigure}[b]{0.5\textwidth}
		\includegraphics[width=\textwidth]{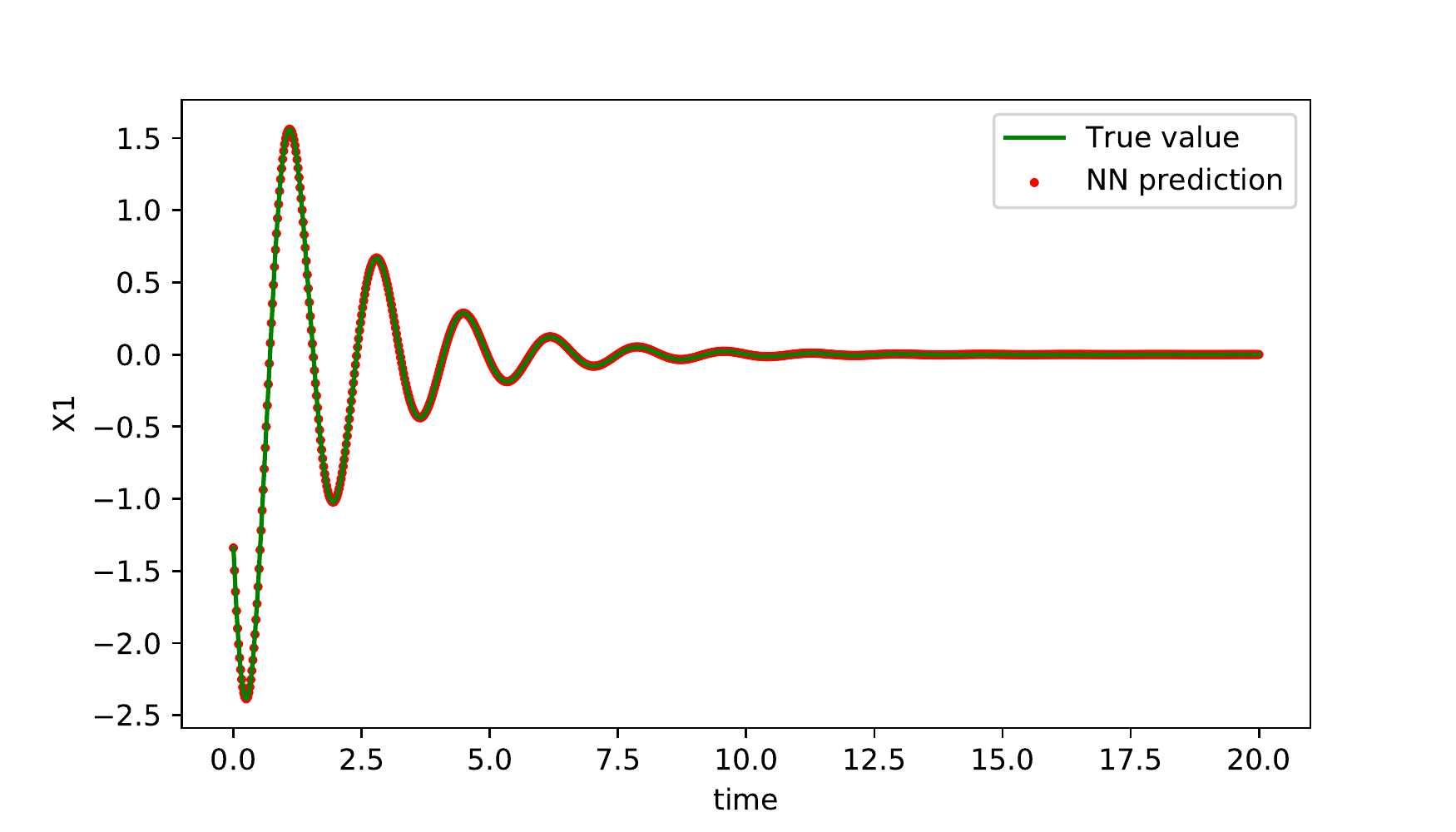}
		\caption{Initial value = $(-1.34,1.7)$.}
		\label{fig:e11_2}
	\end{subfigure}%
    \hfill
	\begin{subfigure}[b]{0.5\textwidth}
		\includegraphics[width=\textwidth]{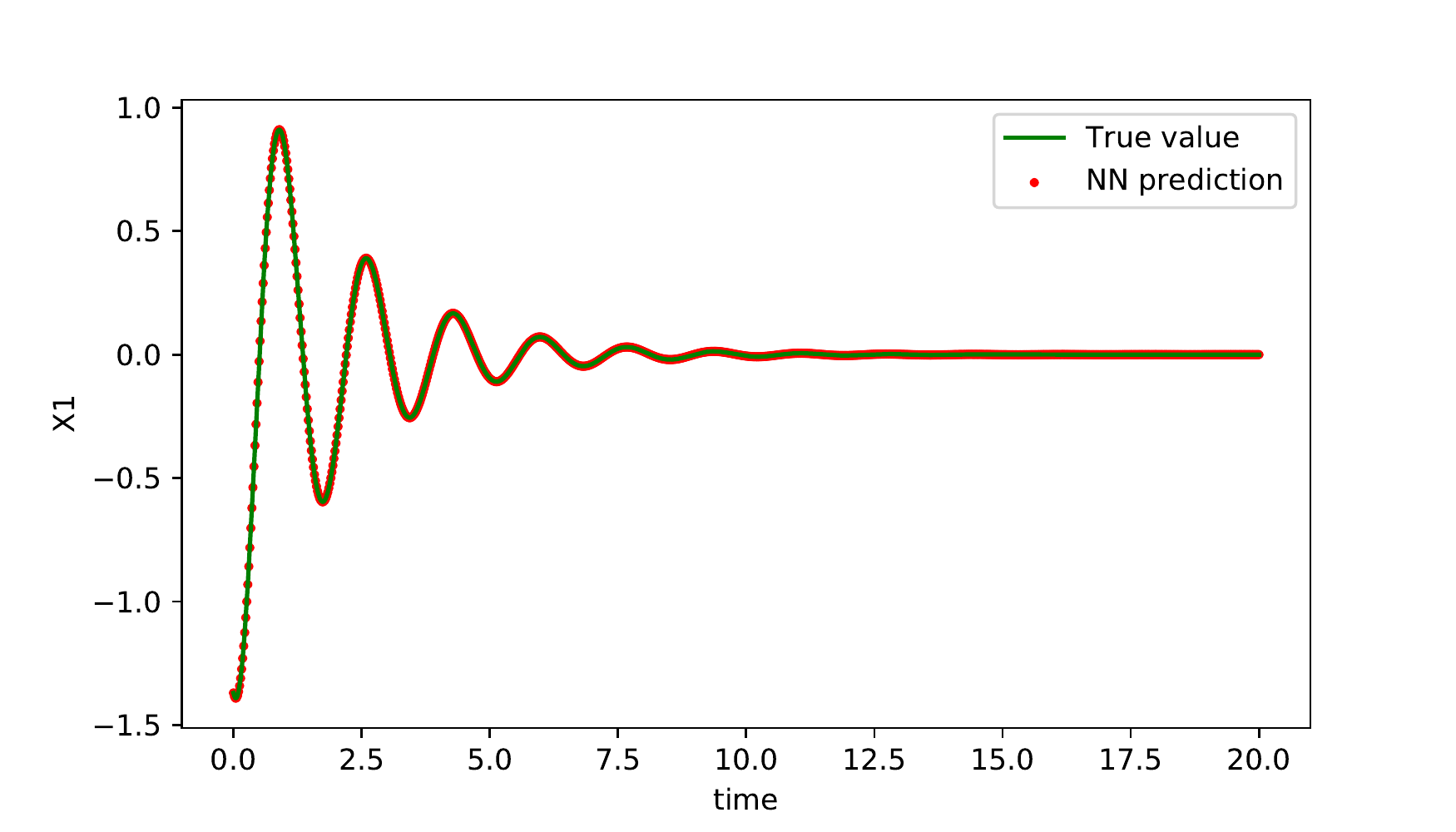}
		\caption{Initial value = $(-1.37,-0.11) $.}
		\label{fig:e11_3}
	\end{subfigure}%
	\begin{subfigure}[b]{0.5\textwidth}
		\includegraphics[width=\textwidth]{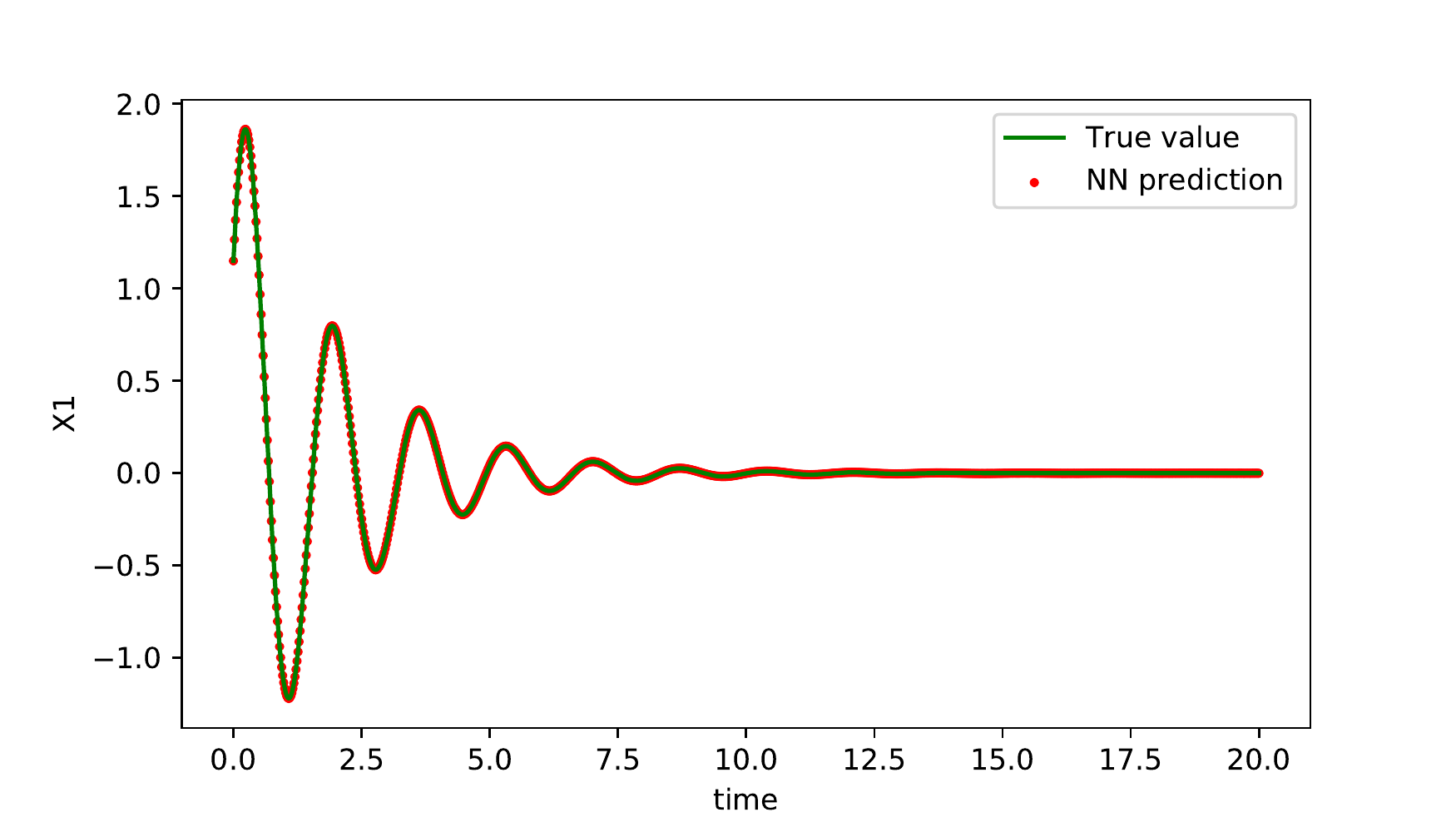}
		\caption{Initial value = $(1.15,-1.21) $.}
		\label{fig:e11_4}
	\end{subfigure}%
    \hfill
	\caption{Example 1: fast decay case. Neural network model
          prediction of $x_1$ with memory length
          $T_M=0.6$ using four different initial conditions.}
	\label{fig:e11_prediction}
\end{figure}

We first consider the fast decay case. In
Fig.~\ref{fig:e11_prediction}, we plot the neural network model
prediction of the observed variable $x_1$ for up to $t=20$, using
memory term $n_M=30$, which corresponds to memory length $T_M=n_M
\Delta = 0.6$. We observe that, for four arbitrarily chosen initial
conditions, the network predictions match the exact true solution
well. Since the solution decay to zero fast, we did not conduct model
prediction over longer term.

\begin{figure}
	\centering
	\begin{subfigure}[b]{\textwidth}
		\includegraphics[width=\textwidth]{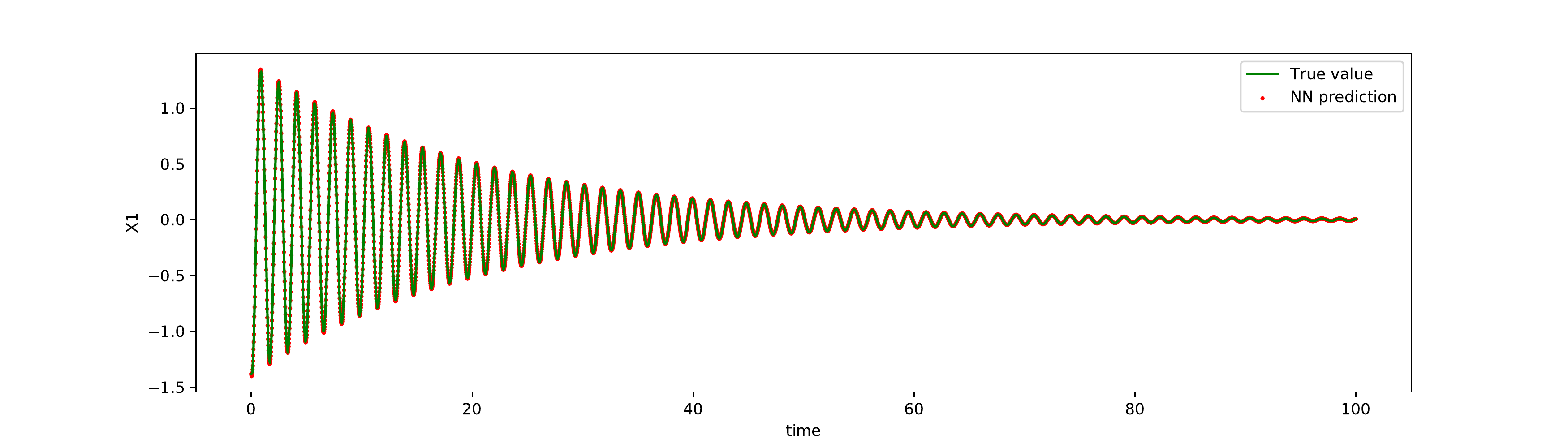}
		\caption{Initial value = $(-1.38,-0.11) $.}
		\label{fig:e12_1}
	\end{subfigure}%
    \hfill
	\begin{subfigure}[b]{\textwidth}
		\includegraphics[width=\textwidth]{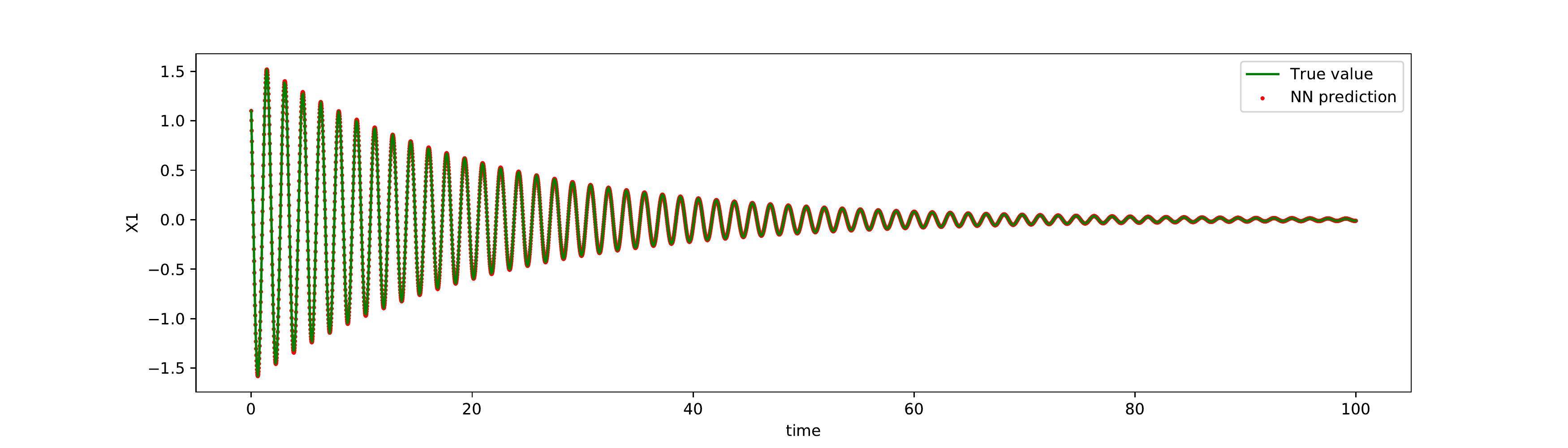}
		\caption{Initial value = $ (1.1,1.45)$.}
		\label{fig:e12_2}
	\end{subfigure}%
	\caption{Example 1: slow decay case. Neural network model
          prediction of $x_1$ with memory length
          $T_M=0.6$ using two different initial conditions.}
        \label{fig:e12_prediction}
      \end{figure}

Longer-term predictions are conducted for the slow decaying case with
$\alpha=1.1$, as shown in Fig.~\ref{fig:e12_prediction}. The NN model
is constructed using memory lenght $T_M=0.6$. The model prediction again shows
good agreement with the exact solution for up to $t=100$.
%%%%%%%%%%%%%%%%%%%%%%%%%%%
%%%%%%%%%%%%%%%%%%%
%%%%%%%%%%%%%%%%%%%%

\begin{figure}
	\centering
	\begin{subfigure}[b]{0.45\textwidth}
		\includegraphics[width=\textwidth]{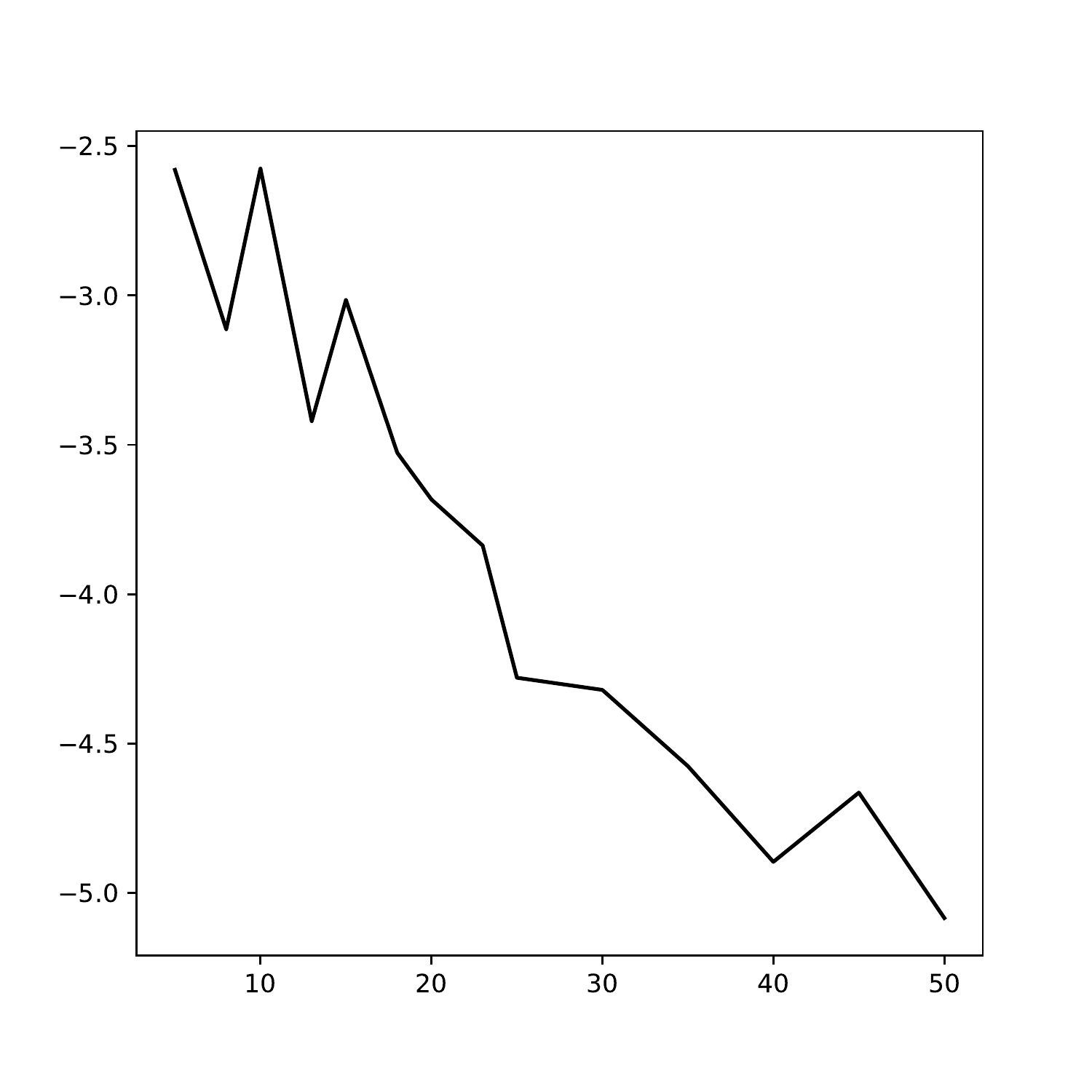}
		\caption{Fast decay case.}
		\label{fig:e11_error_P}
	\end{subfigure}%
	\begin{subfigure}[b]{0.45\textwidth}
		\includegraphics[width=\textwidth]{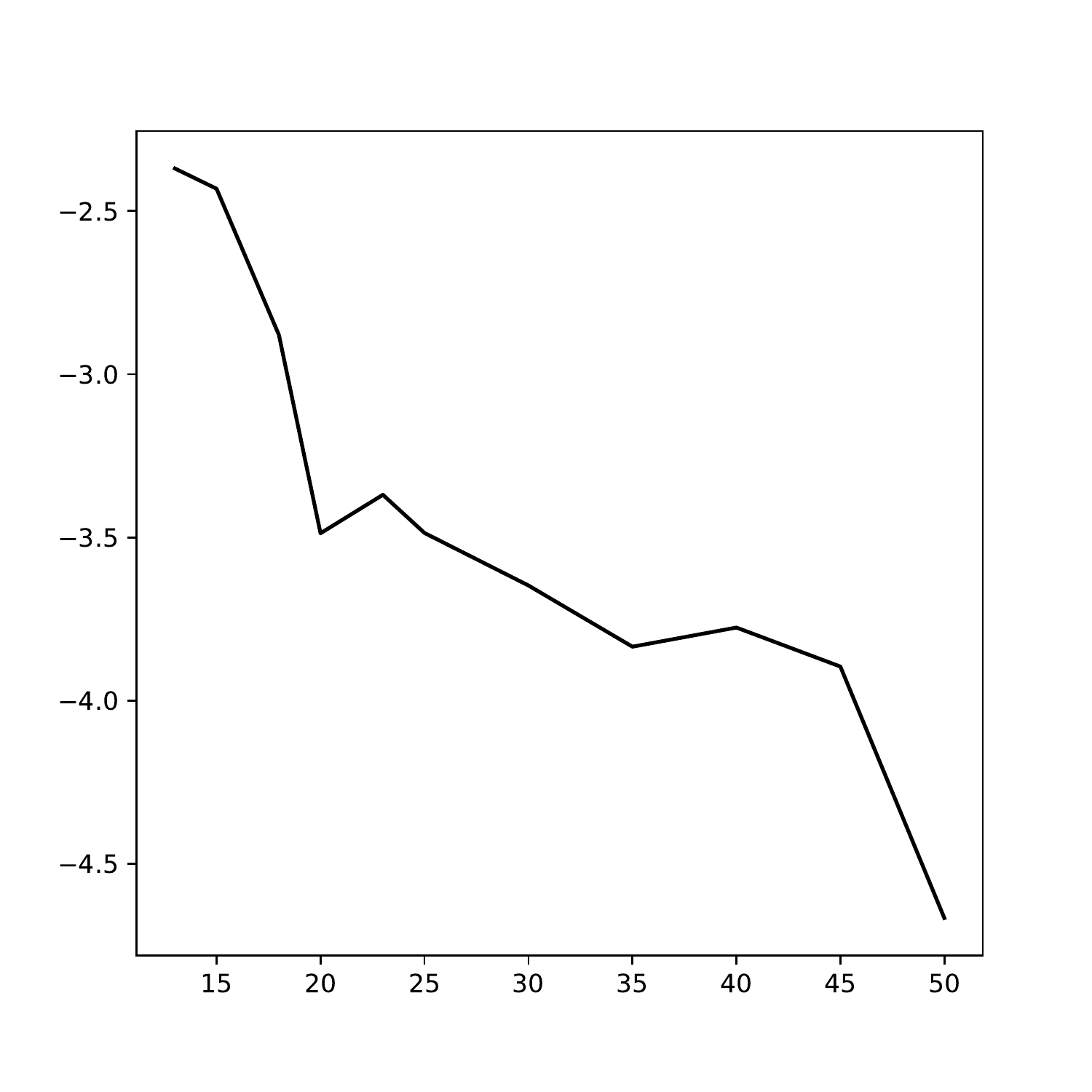}
		\caption{Slow decay case.}
		\label{fig:e12_error_P}
	\end{subfigure}%
	\caption{Example 1: NN prediction errors in $\log$ scale {\em vs.} number of memory terms.}
	\label{fig:error1_P}
\end{figure}

We then examine the effect of different memory length in the NN
modeling. The prediction errors produced by the models with varying $n_M$
are shown in Fig.~\ref{fig:error1_P}. We observe that the NN models
become more accurate as the memory length increases. The accuracy
starts to saturate around $n_M=30$, which corresponds to
memeory length $T_M=0.6$.  Further increasing the
memory length induces no further accuracy improvement.

\subsection{Example 2: Nonlinear System}

\begin{figure}
  \centering
	\includegraphics[width=0.5\textwidth]{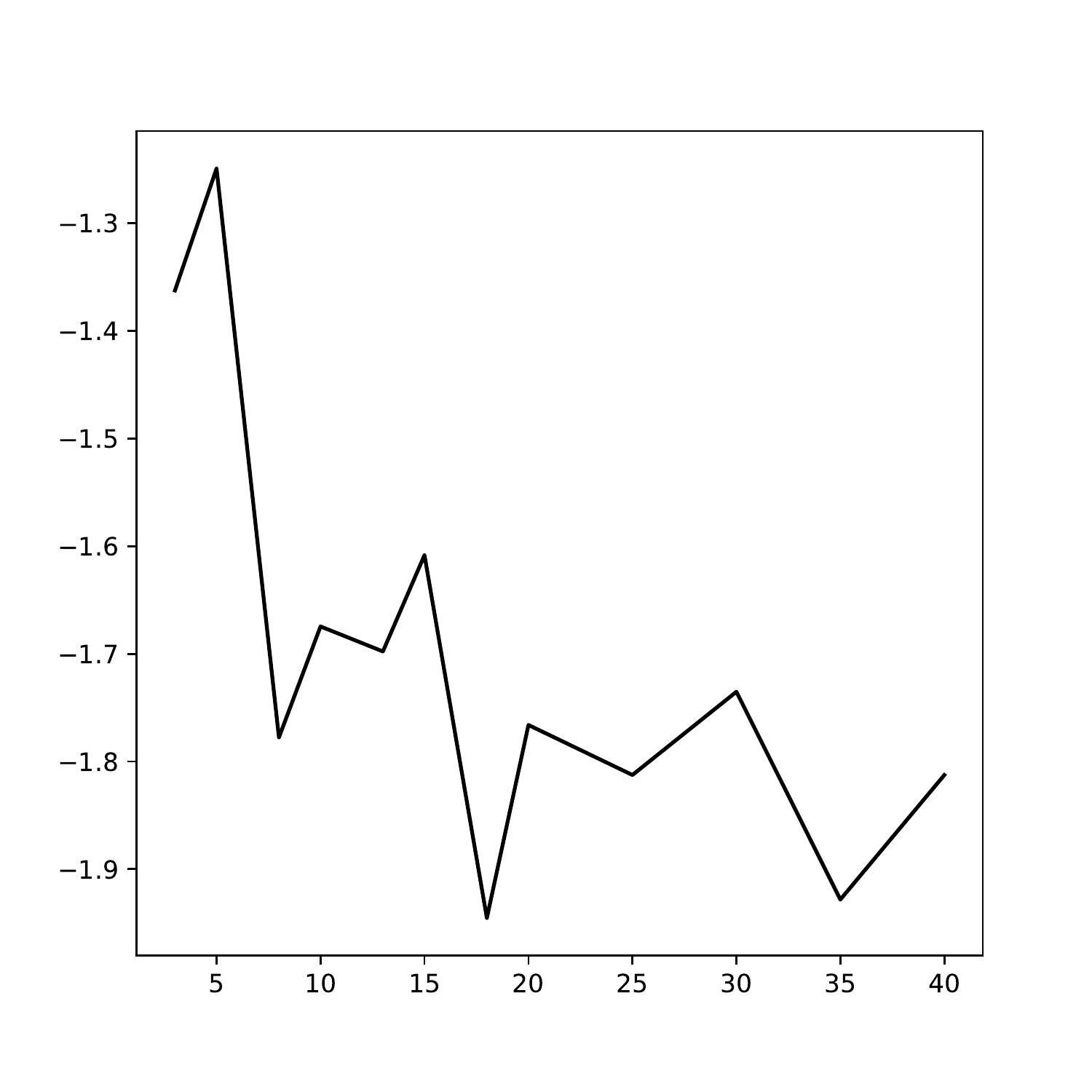}
	\caption{Example 2. Prediction errors in $\log$ scale {\em vs.} the number of
          memory steps.}
	\label{fig:e2_error_P}
\end{figure}

\begin{figure}
	\centering
	\begin{subfigure}[b]{\textwidth}
		\includegraphics[width=\textwidth]{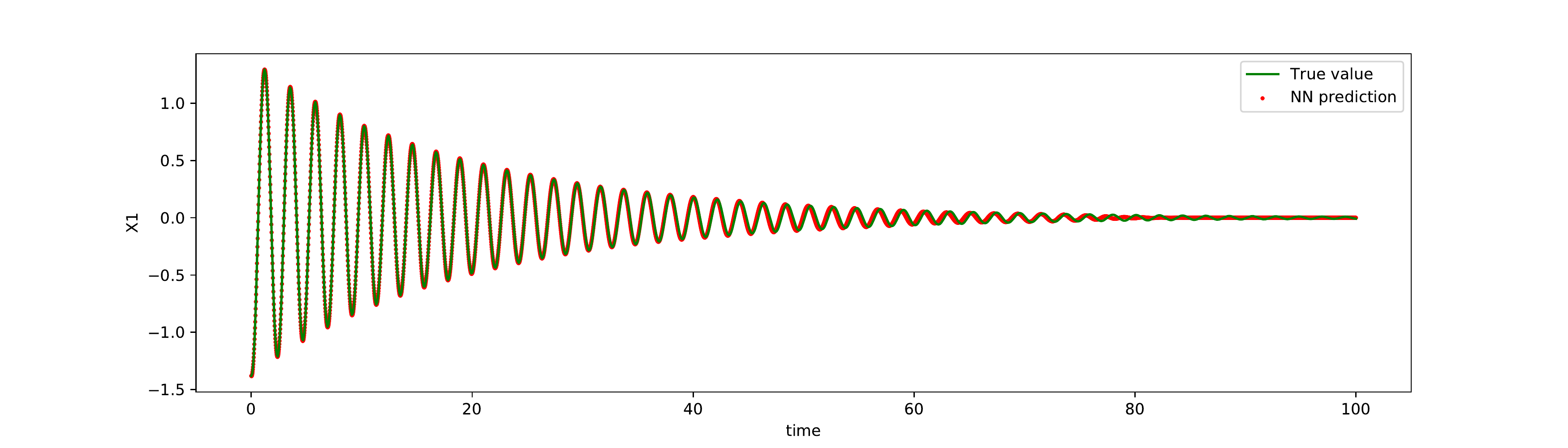}
		\caption{Initial value = $(-1.38,-0.22)$.}
		\label{fig:e2_1}
	\end{subfigure}%
    \hfill
	\begin{subfigure}[b]{\textwidth}
		\includegraphics[width=\textwidth]{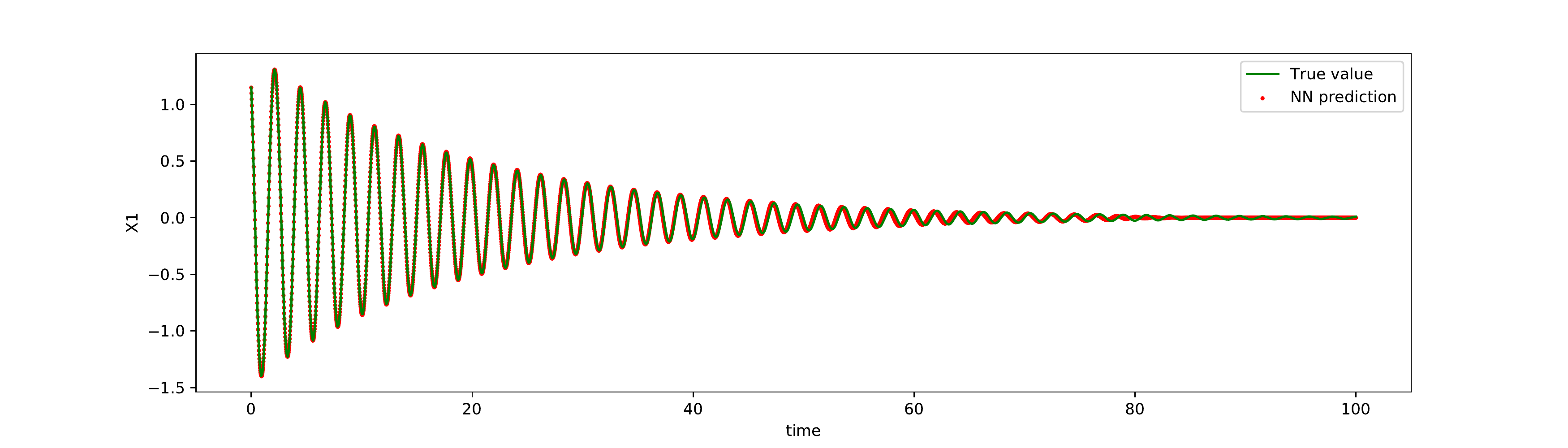}
		\caption{Initial value = $(1.15,-2.43)$.}
		\label{fig:e2_2}
	\end{subfigure}%
	\caption{Example 2. NN model prediction of $x_1$ with memory
          length $T_M=0.4$ using two arbitrary initial conditions.}
	\label{fig:e2_prediction}
\end{figure}

We now consider a damped pendulum system, which is a simple nonlinear system.
\begin{equation}
\begin{cases}
\dot x_1 = x_2,\\
\dot x_2 = -\alpha x_2 - \beta \sin x_1,
\end{cases}       
\end{equation}
where $\alpha=0.1$ and $\beta=8.91$. The domain of interest is set as
$D_{\x} = [-2, 2]\times [-4, 4]$. The observed variable is $z=x_1$ and
we seek to construct an NN model for its prediction. Our training data
sets \eqref{data_set} are generated by collecting $J_0=5$ sequences of $x_1$ data
randomly from each $x_1$ trajectory data of length $K=50$.
The memory step is tested for $n_M = 3, 5, 8, 10, 13, 15, 18, 20$. The
numerical errors in the model predictions with different memory steps
are shown in Fig.~\ref{fig:e2_error_P}. We observe that the accuracy
improvement over increasing $n_M$ starts to saturate with $n_M\geq 10$. The
NN model predictive result with $n_M=20$ is shown in
Fig.~\ref{fig:e2_prediction}. This corresponds to memory length $T_M =
n_M\times \Delta = 0.4$. We again observe very good agreement with the
reference solution for long-term integration up to $t=100$.
%%%%%%%%%%%%%%%%%%%%%

\subsection{Example 3: Chaotic System}

We now consider a nonlinear chaotic system (\cite{PavliotisStuart_2008})
\begin{equation} \label{chaotic}
\begin{cases}
\dot x_1 = -x_2 - x_3,\\
\dot x_2 = x_1 + \frac{1}{5} x_2,\\
\dot x_3 = \frac{1}{5} + y - 5 x_3,\\
\dot y = -\frac{y}{\epsilon} + \frac{x_1 x_2}{\epsilon},
\end{cases}       
\end{equation}
where $\epsilon>0$ is a small parameter. In this example, we choose
the observed variables to be $\z= (x_1, x_2, x_3)^\top$ and let the fast
variable $y$ be the unobserved variable. Note that for this system,
there exists a homogenized system for the slow variables $(x_1, x_2,
x_3)$,
\begin{equation} \label{reduced}
\begin{cases}
\dot x_1 = -x_2 - x_3,\\
\dot x_2 = x_1 + \frac{1}{5} x_2,\\
\dot x_3 = \frac{1}{5} + x_3(x_1-5).
\end{cases}       
\end{equation}
The reduced system is a good approximation for the true system when
$\epsilon \ll 1$. Here we will construct NN models for the reduced
variables $\z$ and compare the prediction results against the true
solution of \eqref{chaotic}, as well as those obtained by the reduced
system \eqref{reduced}. We set $\epsilon=0.01$, in which case the
reduced system \eqref{reduced} is considered an accurate approximation of the true system.

The domain of interest is set to be
$D_\x=[-7.5,10]\times[-10,7.5]\times[0,18]\times[-1,100]$, which is
sufficiently large to enclose the solution trajectories for different
initial conditions.
%But in addition, I purposely add more data points around 0 for $x_3$
%and $y$. So only a proportion of 0.6 of the data comes from $D$, the
%remaining proportion of 0.4 comes from
%$D^{+}=[-7.5,10]\times[-10,7.5]\times[0,0.5]\times[-0.15,2.5]$.
To generate training data set \eqref{data_set}, we solve the true
system \eqref{chaotic} using randomly sampled initial conditions via a high resolution numerical solver and
record the trajectories of $\z$ with length $K=100$, which corresponds
to a time lapse of $K\Delta = 2$. In each trajectory, we randomly
select $J_0=5$ sequences of data with length $(n_M+2)$ for our
training data set.
Different memory steps of $n_M=10,20,30,40,50,60,70,80$ are
examined. This corresponds to memory length $T$ ranging from 0.2 to
1.6.
Our results indicate that $n_M=60$, i.e, $T_M=1.2$, delivers accurate
predictions. Further increasing memory length does not lead to better predictions.

\begin{figure}
	\centering
	\includegraphics[width=\textwidth]{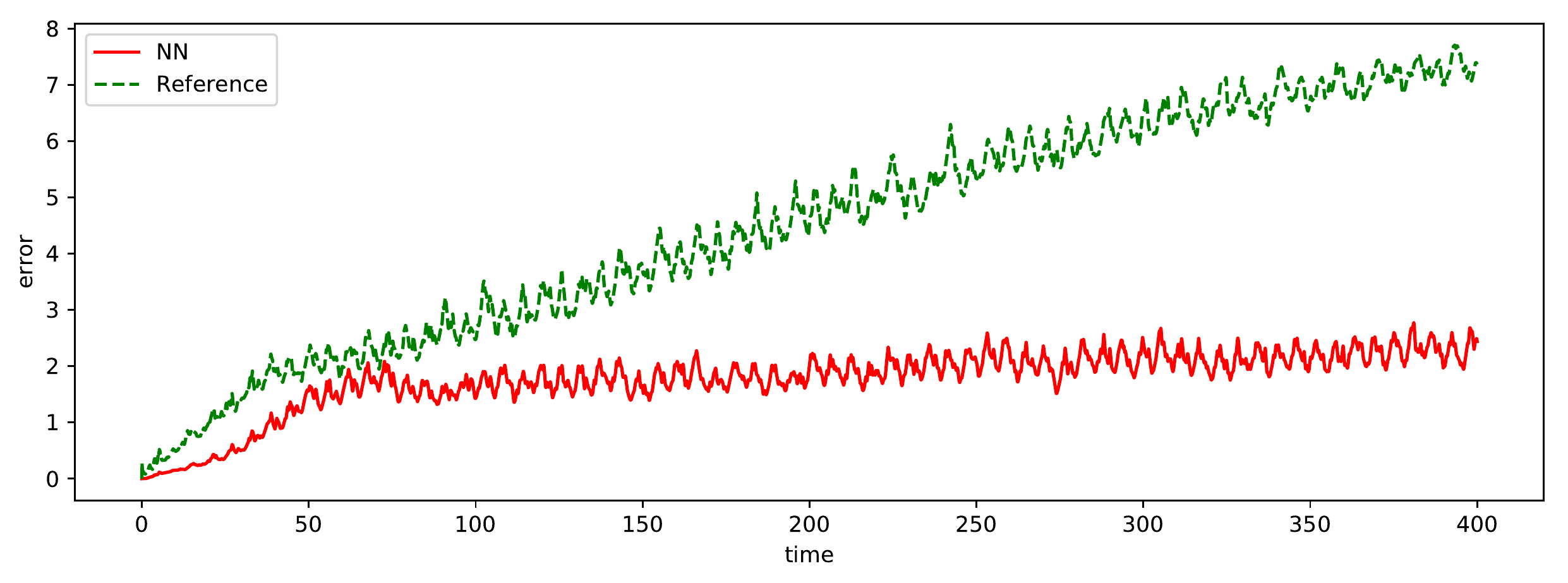}
	\caption{Example 3. Errors in model predictions over time, by the NN model
          with $T_M=1.2$ and the reduced system
          \eqref{reduced}. The errors are averaged over 100
          simulations of different initial conditions.}
	\label{fig:e3_ave}
\end{figure}

\begin{figure}
	\centering
	\begin{subfigure}[b]{\textwidth}
		\includegraphics[width=\textwidth]{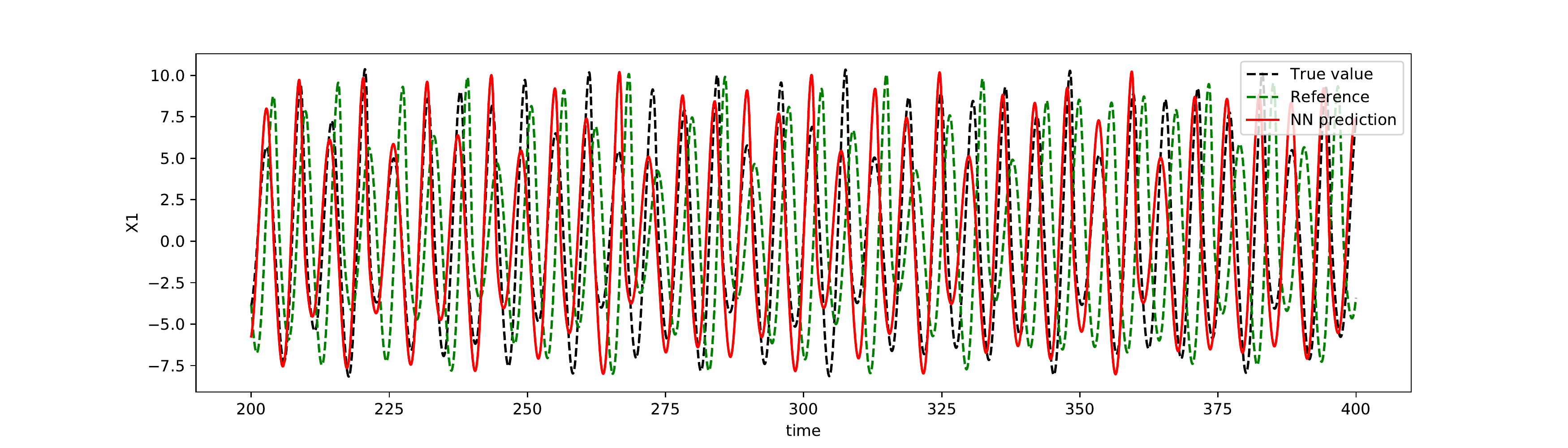}
		%\caption{$x1$}
		%\label{fig:e3_1}
	\end{subfigure}%
	\hfill
	\begin{subfigure}[b]{\textwidth}
		\includegraphics[width=\textwidth]{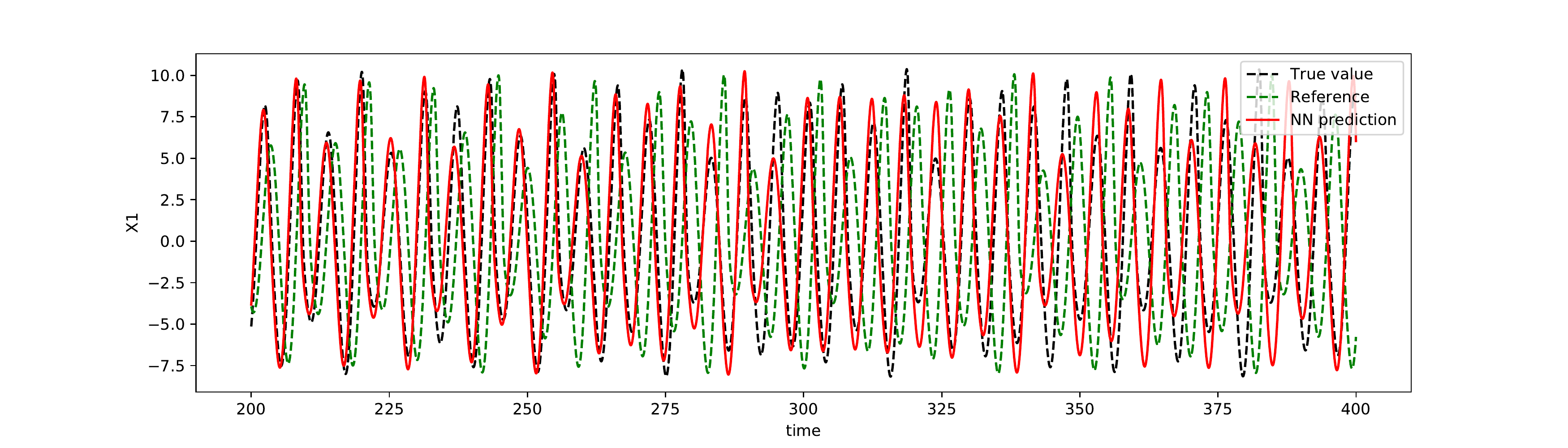}
		%\caption{x1, initial value = }
		%\label{fig:e3_2}
	\end{subfigure}%
    \hfill
    \begin{subfigure}[b]{\textwidth}
    	\includegraphics[width=\textwidth]{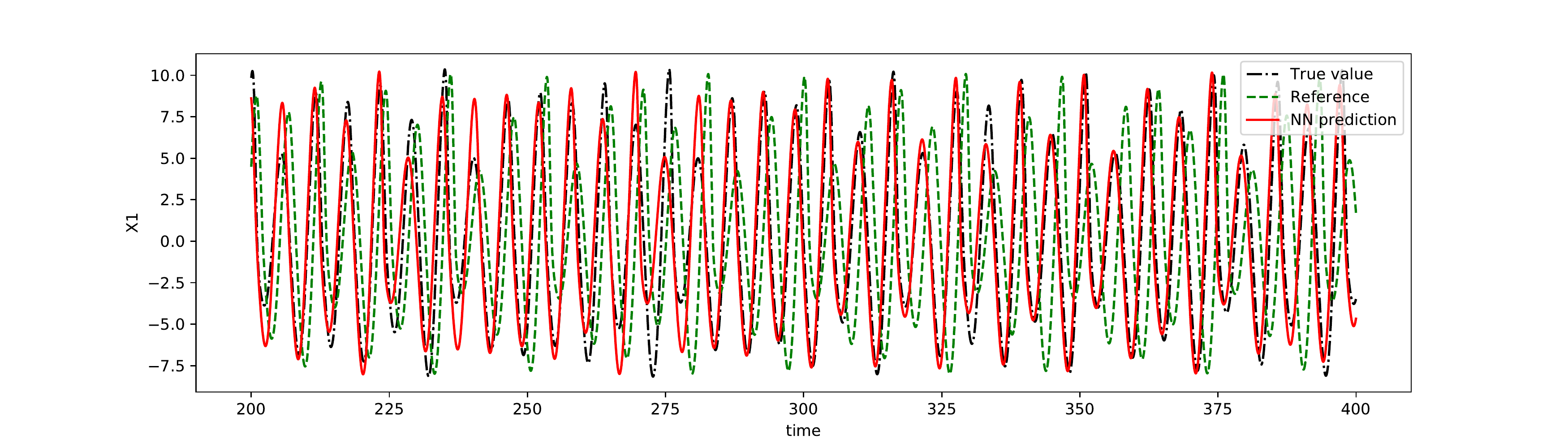}
    	%\caption{x1, initial value = }
    	%\label{fig:e3_3}
    \end{subfigure}
	\caption{Example 3. Long-term model prediction of $x_1$ by the NN model
        with $T_M=1.2$ and the reduced system \eqref{reduced}. Zoomed
        view for $t\in [200, 400]$ with three aribitrarily chosen
        initial conditions.}
	\label{fig:e3_prediction}
\end{figure}

The evolution of prediction errors, measured in $\ell_2$-norm against the reference
true solution, are shown in
Fig.~\ref{fig:e3_ave}, for both our neural network model and the
reduced model \eqref{reduced} for long term prediction up to $t=400$.
Here the errors are averaged over 100 simulations using randomly
selected initial conditions. We
observe that our NN model produces noticebaly more accurate results than the
reduced system \eqref{reduced}, even for the case of small $\epsilon$
when \eqref{reduced} is supposed to be highly accurate. More
importantly, the NN model exhibits much smaller error growth over long
time, compared to the reduced system \eqref{reduced}.
The solution
behavior is shown in Fig.~\ref{fig:e3_prediction}, for long-term
integration up to $t=400$ for the first component $x_1$. (Behavior for
other components are similar.) The NN model produces visually better
results than the reduced system \eqref{reduced}, when compared to the
true solution, especially in term of capturing the phase/frequency of
the solution.
%%%%%%%%%%%%%%%%%%%%%

\subsection{Example 4: Larger Linear System} \label{sec:Example4}

We now consider a larger linear system involving 20 state variables
$\x=(\p; \q)\in\Rs^{20}$,
\begin{equation}
\begin{cases}
\dot \p = \Sigma_{11} \p + (\I + \Sigma_{12}) \q,\\
\dot \q = -(\I + \Sigma_{21}) \p - \Sigma_{22} \q,
\end{cases}       
\end{equation}
where $\p, \q \in \Rs^{10}$, $\I$ is identity matrix of size $10\times
10$, and $\Sigma_{ij} \in \Rs^{10\times 10}$,
$i=1,2,j=1,2$.
%Entries in $\Delta_{ij}$ are assumed to be small and
%treated as disturbance.
We set the observed variables to be $\z=\p\in\Rs^{10}$ and let $\w=\q\in\Rs^{10}$ be the
unobserved variables.
Note that as a linear system \eqref{general},  the exact Mori-Zwanzig equations for the reduced variables
$\z$ are available in analytical form \eqref{linearMZ}.
We set the entries of $\Sigma_{ij}$ to be small and consider them as
perturbations to an oscillatory system.
The exact values of the entries of matrices $\Sigma_{ij}$ are presented in
Appendix, for self completeness of the paper.

\begin{figure}
	\centering
	\begin{subfigure}[b]{0.5\textwidth}
		\includegraphics[width=\textwidth]{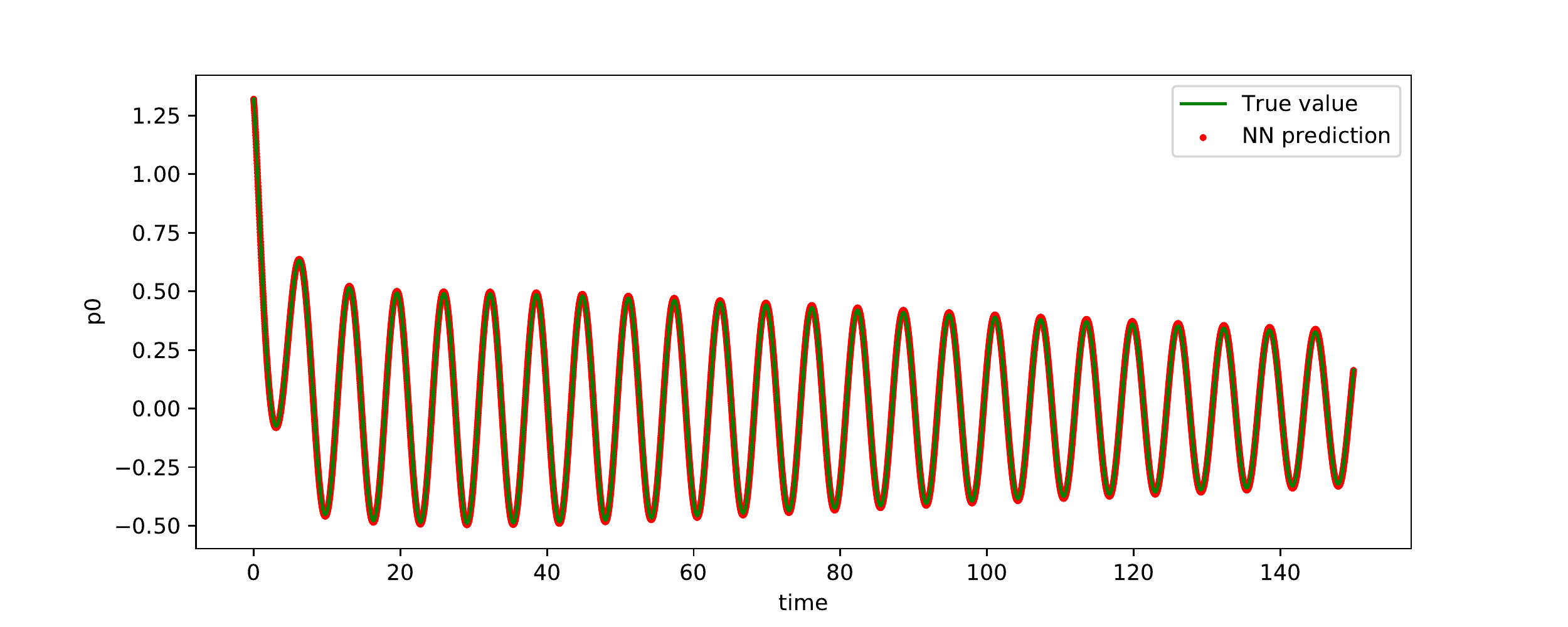}
		%\caption{x1, initial value = }
	\end{subfigure}%
	%~ %add desired spacing between images, e. g. ~, \quad, \qquad, \hfill etc. 
	\begin{subfigure}[b]{0.5\textwidth}
		\includegraphics[width=\textwidth]{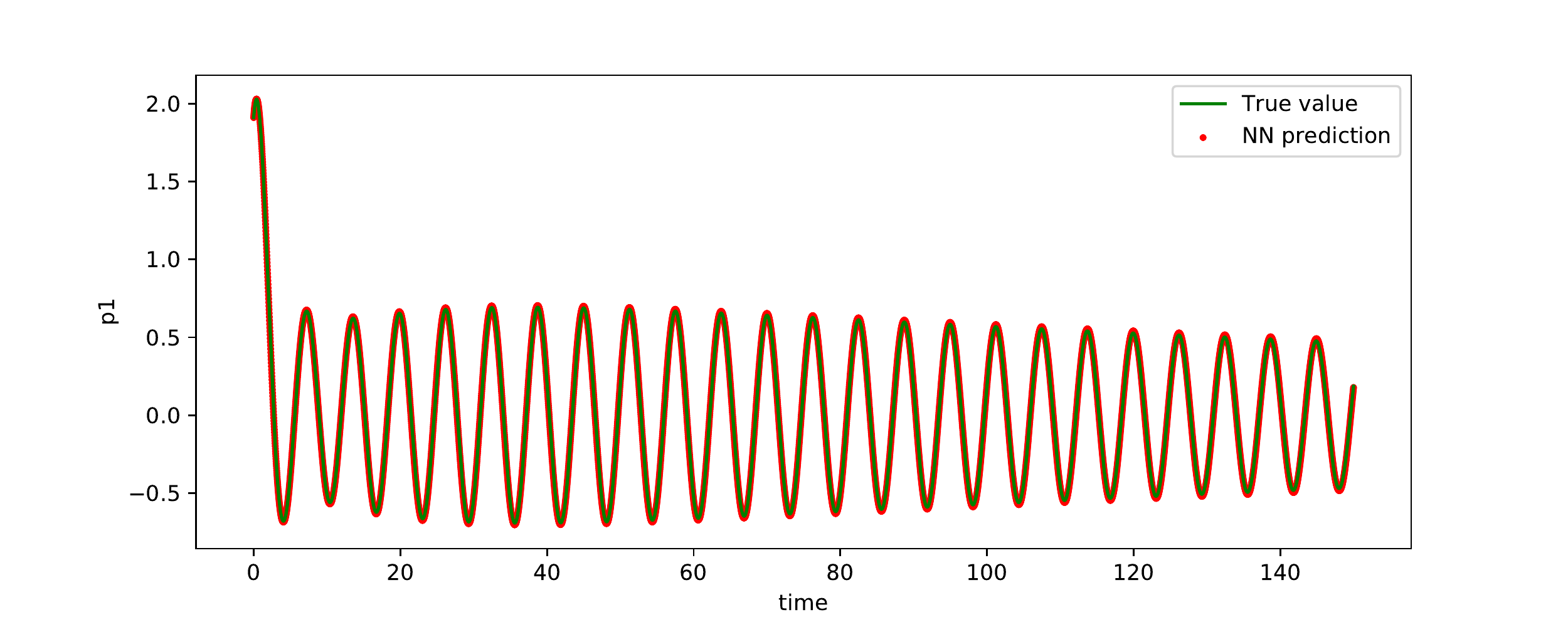}
		%\caption{x1, initial value = }
	\end{subfigure}%
	\hfill
	\begin{subfigure}[b]{0.5\textwidth}
		\includegraphics[width=\textwidth]{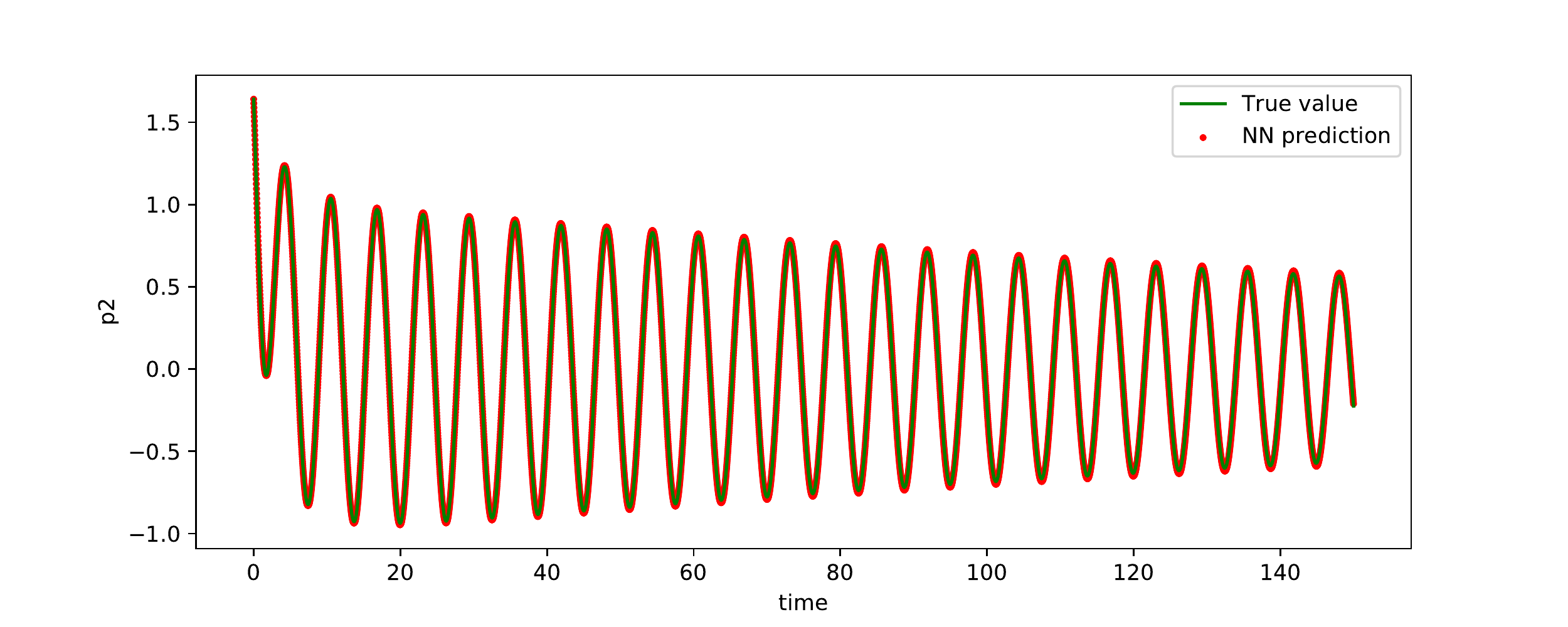}
		%\caption{x1, initial value = }
	\end{subfigure}%
	\begin{subfigure}[b]{0.5\textwidth}
		\includegraphics[width=\textwidth]{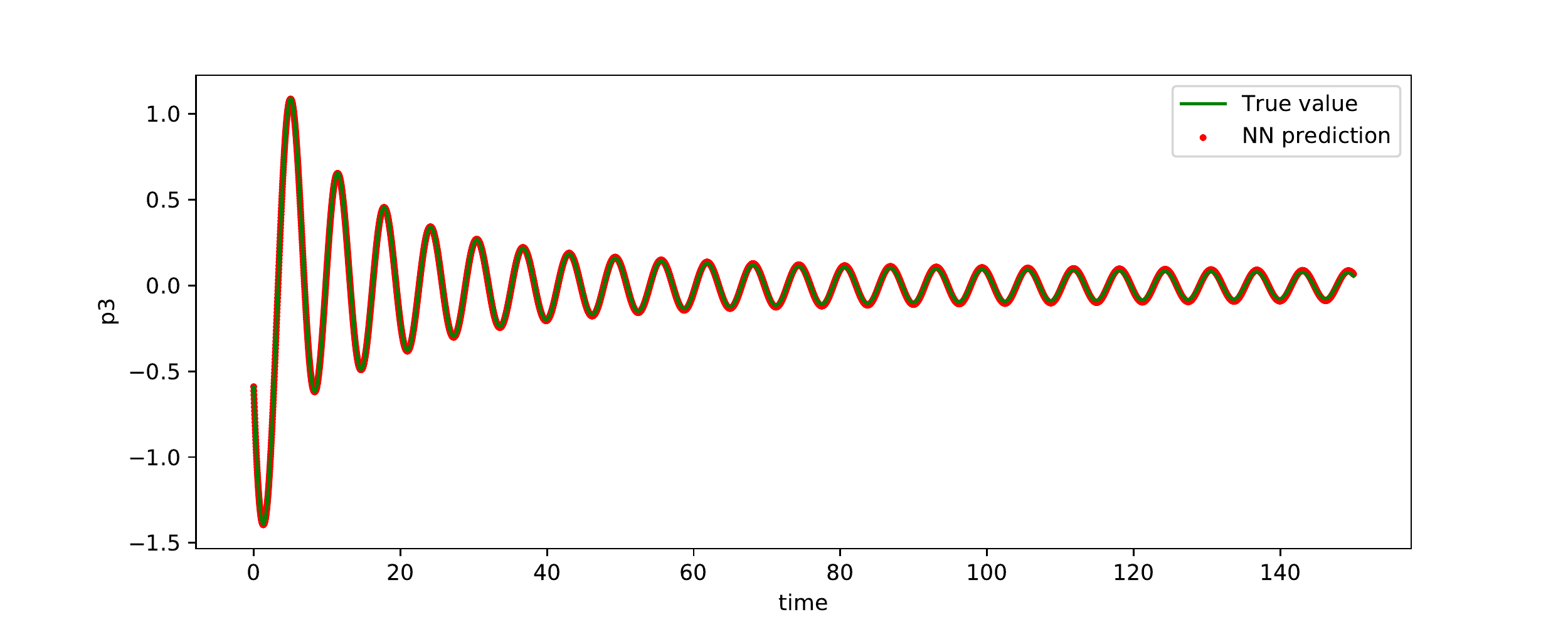}
		%\caption{x1, initial value = }
	\end{subfigure}%
    \hfill
    \begin{subfigure}[b]{0.5\textwidth}
    	\includegraphics[width=\textwidth]{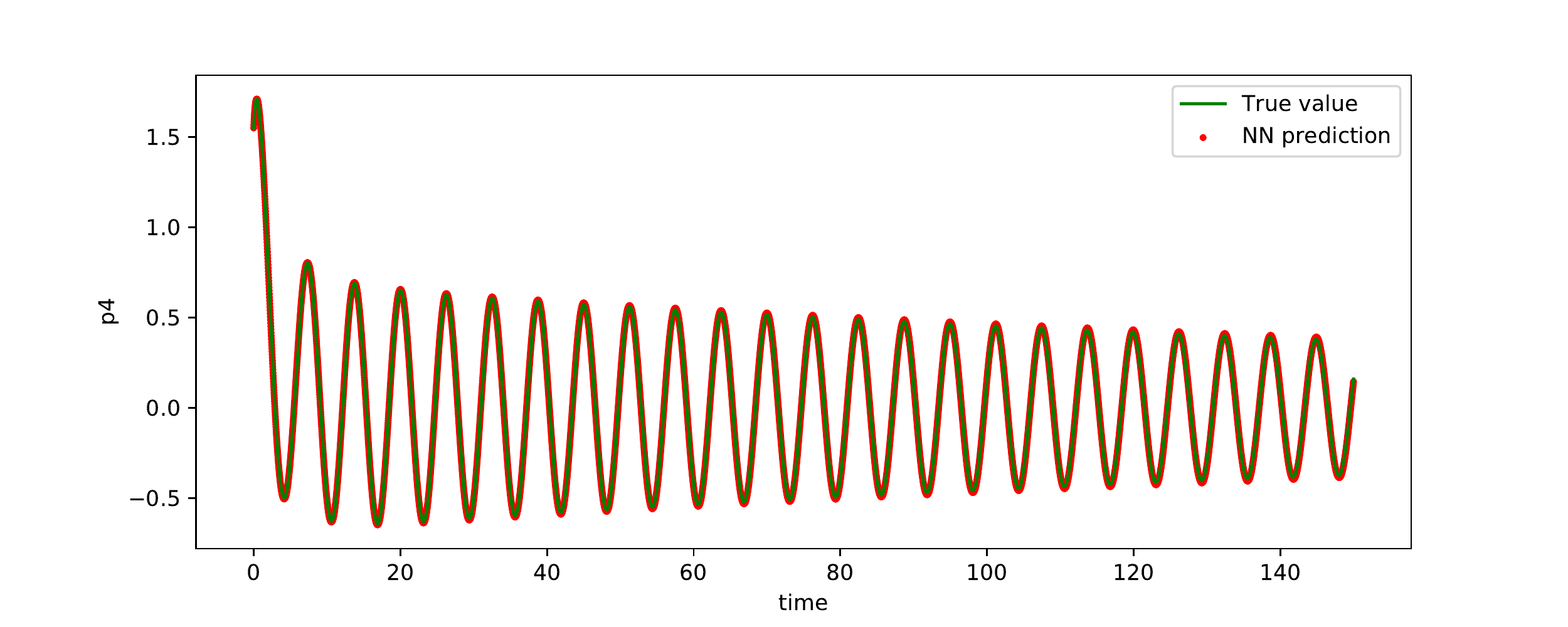}
    	%\caption{x1, initial value = }
    \end{subfigure}%
    \begin{subfigure}[b]{0.5\textwidth}
    	\includegraphics[width=\textwidth]{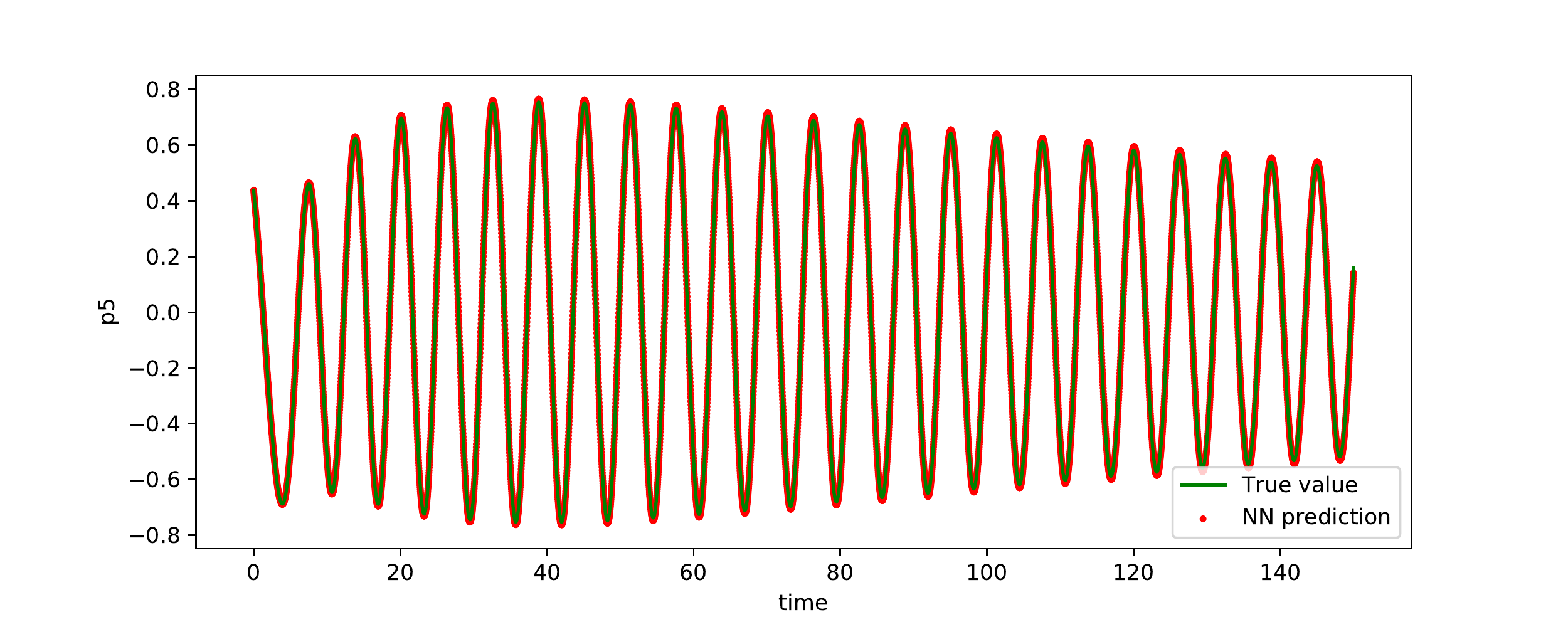}
    	%\caption{x1, initial value = }
    \end{subfigure}%
    \hfill
    \begin{subfigure}[b]{0.5\textwidth}
    	\includegraphics[width=\textwidth]{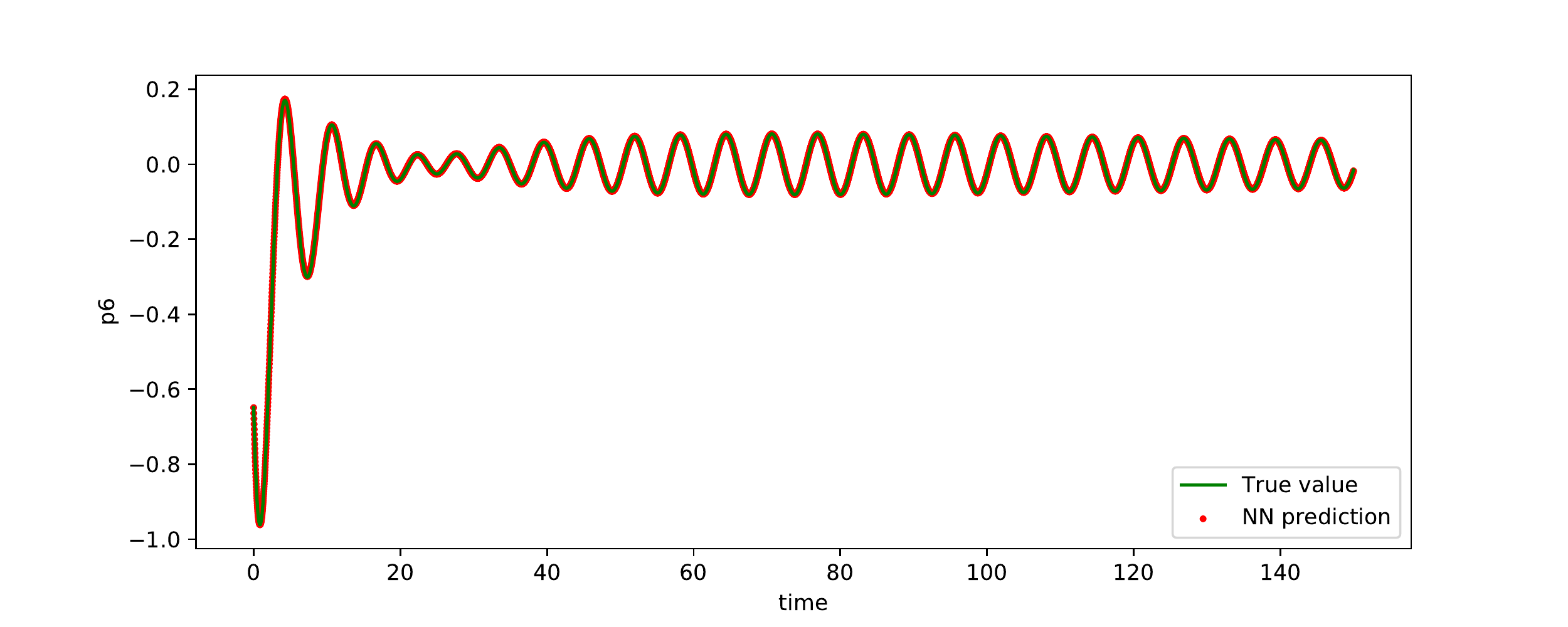}
    	%\caption{x1, initial value = }
    \end{subfigure}%
    \begin{subfigure}[b]{0.5\textwidth}
    	\includegraphics[width=\textwidth]{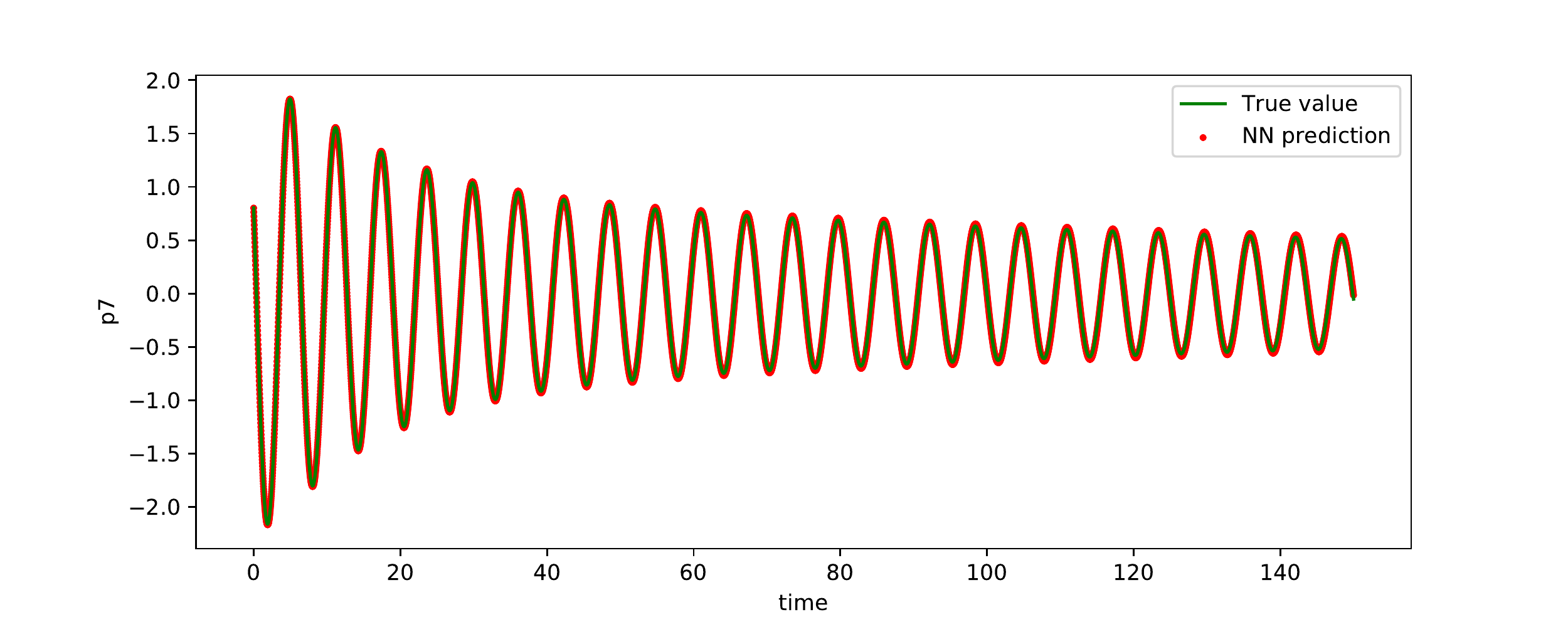}
    	%\caption{x1, initial value = }
    \end{subfigure}%
    \hfill
    \begin{subfigure}[b]{0.5\textwidth}
    	\includegraphics[width=\textwidth]{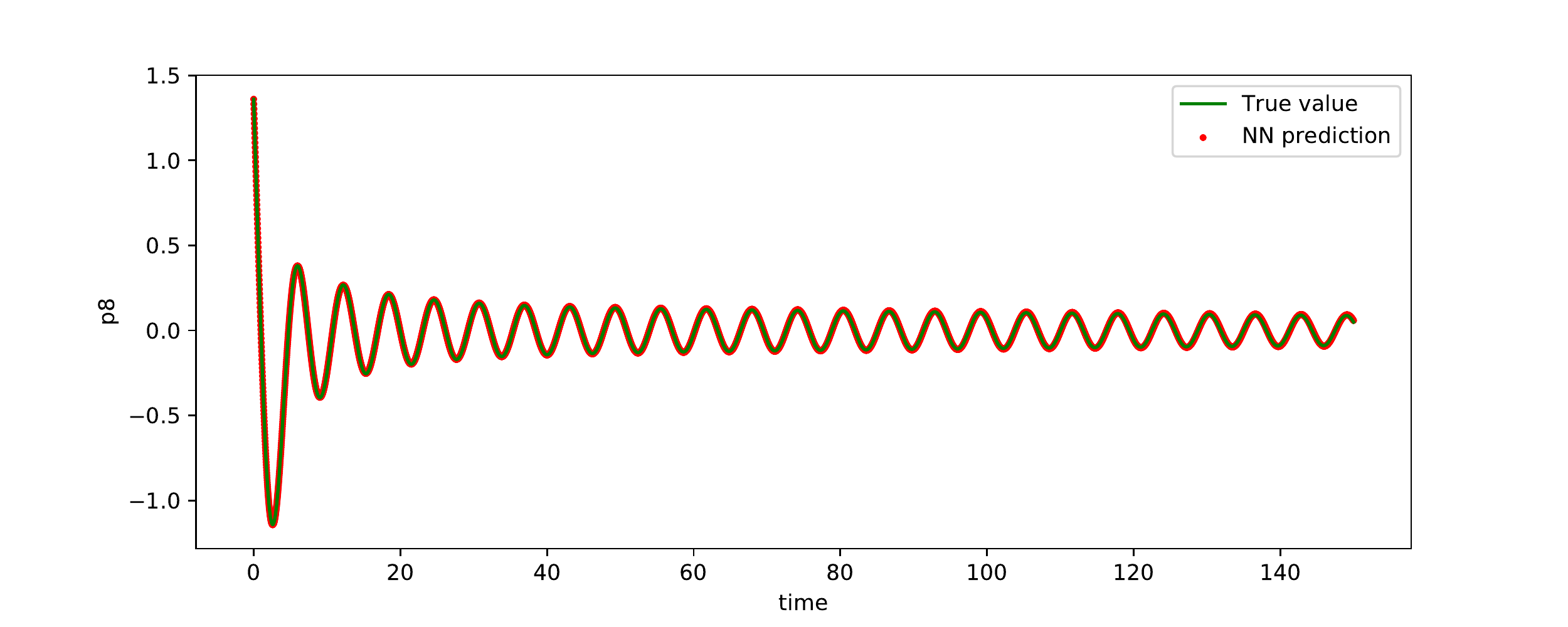}
    	%\caption{x1, initial value = }
    \end{subfigure}%
    \begin{subfigure}[b]{0.5\textwidth}
    	\includegraphics[width=\textwidth]{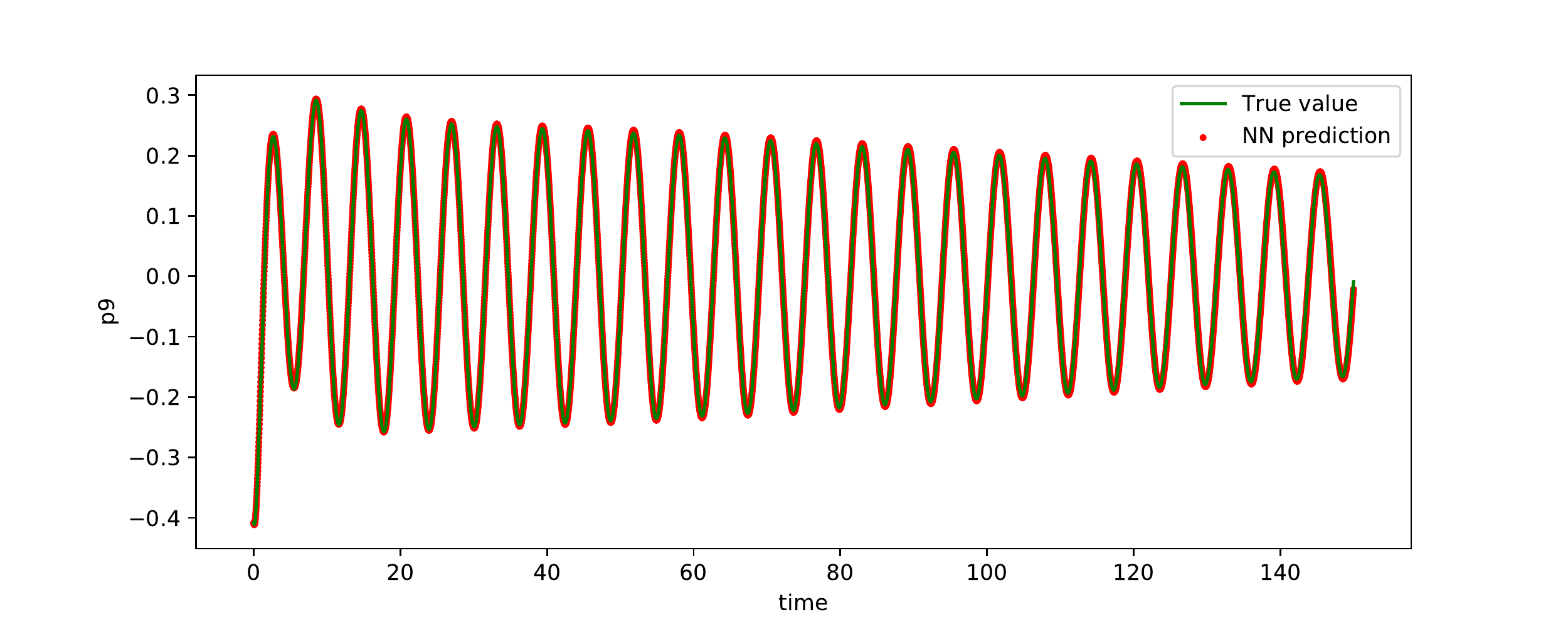}
    	%\caption{x1, initial value = }
    \end{subfigure}%
	\hfill
	\caption{Example 4. NN model prediction of $\p$ for up to
          $t=150$, with memory length $T_M=0.6$.}
	\label{fig:e4_prediction}
\end{figure}

The domain of interest is set to be $D_{\x}=[-2,2]^{20}$. In
generating the training data sets, we randomly select $J_0=5$
sequences of data from each trajectory of $\z$ with length $K=100$. We test
different memory steps for $n_M=10,15,20,25,30,35,40,45,50$. Our
experiments show that $n_M=30$, which corresponds to a memory length
$T_M=0.6$, provides accurate prediction. Further increasing memory
length does not lead to better prediction accuracy.
Our NN model predictions for long-term integration up to $t=150$ are
shown in Fig.~\ref{fig:e4_prediction}. Compared to the reference
solution obtained from the true system, we observe very good
agreement, where the NN predictions overlap the true solutions to be
visually indistinguishable.

%\begin{frame}
%	
%	\resizebox{0.95\linewidth}{!}{%
%		$\displaystyle
%		\left( \begin{array}{rrrrrrrrrr} 
%		-0.00609 & 0.00579 & -0.000945 & -0.0121 &0.00938 & -0.0124 & 0.00492 & 0.00371 & 0.00117 & 0.00473\\
%		-0.00609 & 0.00579 & -0.000945 & -0.0121 &0.00938 & -0.0124 & 0.00492 & 0.00371 & 0.00117 & 0.00473\\
%		\end{array} \right)
%		$}
%\end{frame}

\section{Conclusion} \label{sec:conclusions}

We present construction of deep neural network (DNN) model to
approximating unknown dynamical systems when only a subset of
variables are observed. The DNN model then provides a reduced model
for the unknown dynamical system. 
Based on Mori-Zwanzig (MZ) formulation for reduced systems, we established
a discrete Mori-Zwanzig formulation with finite memory assumption. We
then designed a straightforward DNN structure to explicitly
incorporate the system memory into the predictive model.
Numerical tests on both linear and nonlinear systems demonstrated good
accuracy of the DNN models. This invites further in-depth study of the
approach, both theoretically and numerically.

\bibliographystyle{siamplain}
\bibliography{LearningEqs,neural}

\appendix

\section{Details of Example 4 in Section \ref{sec:Example4}}

The detailed setting of Example 4 is
\begin{equation}
\begin{cases}
\dot \p = \Sigma_{11} \p + (\I + \Sigma_{12}) \q,\\
\dot \q = -(\I + \Sigma_{21}) \p - \Sigma_{22} \q,
\end{cases}       
\end{equation}
where $\p, \q \in \Rs^{10}$, $\I$ is identity matrix of size $10\times
10$, and $\Sigma_{ij} \in \Rs^{10\times 10}$,
$i=1,2,j=1,2$.
The matices $\Sigma_{ij}$ are defined as follows.

%%%%%%%%%%%%%%%%%%%
\begin{frame}
	\footnotesize
	\setlength{\arraycolsep}{3.2pt} % default: 5pt
	\begin{align*} 
	&\Sigma_{11}\times 10^3 = \\
	&
	\left( \begin{array}{rrrrrrrrrr} 
	-6.09 & 5.79 & -0.945 & -12.1 &9.38 & -12.4 & 4.92 & 3.71 & 1.17 & 4.73\\
	-9.57 & -8.88 & -12.1 & -12.9 & 5.11 & 26.5 & -7.33 & -8.01 & -21.6 & -10.2\\
	-0.733 & 6.2 & 10.7 & -6.06 & -7.07 & -1.7 & -16.4 & 6.69 & -1.59 & 7.69\\
	7.83 & 12.5 & 5.77 & -14.9 & -17.8 & -1.01 & -4.05 & -15 & -6.61 & -4.94\\
	8.1 & 4.13 & 4.21 & 23.3 & -4.63 & 1.77 & -14.9 & 17.9 & -17.1 & -8.19\\
	-7.68 & 6.98 & 27.6 & 19 & 20.9 & 12.2 & 15.6 & -11.2 & -3.56 & -2.47\\
	-14.9 & -5.73 & -19.7 & -8.77 & -9.17 & -2.95 & -9.48 & -2.95 & 5.43 & 15.4\\
	-1.84 & 2.05 & -1.98 & 3.83 & -4.06 & 7.72 & 4.04 & -13.7 & 20.3 & 0.509\\
	12.1 & 19.7 & -14.3 & 12.6 & -4.67 & 9.72 & 5.87 & 0.664 & -10.8 & -18.2\\
	3.07 & 3.65 & 3.88 & 7.44 & 12.7 & 13.5 & -6.66 & -23.9 & -11.7 & 16.6
	\end{array} \right),
	\end{align*}
\end{frame}

\begin{frame}
	\footnotesize
	\setlength{\arraycolsep}{3pt} % default: 5pt
	\begin{align*} 
		&\Sigma_{12} \times 10^3= \\
		&
		\left( \begin{array}{rrrrrrrrrr} 
			11.7 & -12.3 & -8.87 & -6.86 & -9.6 & 11 & 25.6 & -0.155 & 17.8 & -10.9\\
			12.9 & 3.28 & 2.84 & 3.35 & 16.6 & 5.96 & 6.99 & -20.2 & 8.37 & -8.87\\
			-0.154 & -16.5 & 12.1 & 0.381 & 11.2 & -2.59 & 12.8 & 3.32 & -10.9 & -3.81\\
			6.49 & 15.8 & -0.273 & 9.05 & -3.15 & 0.976 & -7.35 & 0.889 & 6.41 & 15.6\\
			4.86 & -1.52 & 0.118 & 17.8 & -5.08 & -4.96 & -2.89 & 3 & 22.4 & 16.4\\
			7.83 & -9.66 & -2.09 & 5.97 & 3.97 & 19.2 & 4.03 & -15.3 & -8.5 & -15.8\\
			-4.61 & -4.98 & 17 & -14 & -17.5 & 0.104 & -27.5 & 10.9 & -17.9 & -5.9\\
			3.88 & 14 & -2.63 & -7.27 & -21 & -0.403 & -2.18 & -22 & 2.01 & -2.45\\
			14.4 & -4.65 & -8.67 & -23.2 & -2.73 & 9.58 & -13.9 & 0.415 & 10.3 & 17.5\\
			-16.8 & 8.18 & -12.3 & 14.2 & -18.4 & -10.2 & -11.4 & -1.99 & -2.65 & -2.34
		\end{array} \right),
	\end{align*}
\end{frame}

\begin{frame}
	\footnotesize
	\setlength{\arraycolsep}{3.2pt} % default: 5pt
	\begin{align*} 
		&\Sigma_{21}\times 10^3 = \\
		&
		\left( \begin{array}{rrrrrrrrrr} 
			-3.51 & -4.91 & -4.51 & -15.8 & -12 & -5.72 & -9.52 & -14.3 & 0.745 & -11.8\\
			1.8 & 2.07 & 8.78 & 5.3 & -5.25 & 5.7 & 0.0957 & 9.77 & 2.17 & 12.8\\
			-9.87 & 5.19 & 0.884 & 2.59 & -7.95 & 5.56 & 6.41 & 16.4 & 15.6 & 14.3\\
			10.4 & 7.14 & 15.5 & -6.6 & 5.33 & -3.37 & 2.8 & -9.61 & 8 & -16.8\\
			15.5 & 19.6 & -1.1 & 0.6 & 8.38 & 7.62 & 3.43 & 1.28 & 10.3 & -4.76\\
			0.119 & -9.43 & -6.6 & -9.99 & -10.5 & 17.8 & 13.5 & -6.63 & -0.566 & -1.81\\
			-6.77 & -1.42 & 7.46 & 3.32 & 11.7 & 1.3 & -6.21 & 6.9 & 3.89 & 18.9\\
			2.93 & 15.1 & -4.65 & 11.1 & 9.13 & -9.58 & -7.04 & 6.88 & -4.07 & 10.2\\
			-6.02 & 14 & -5.91 & -4.92 & 0.851 & 0.652 & -2.57 & 0.835 & -5.14 & 10.6\\
			1.41 & 5.8 & -2.31 & 6.17 & 13.3 & 3.57 & 15.9 & -0.753 & -0.818 & -10.3
		\end{array} \right),
	\end{align*}
\end{frame}

\begin{frame}
	\footnotesize
	\setlength{\arraycolsep}{4.4pt} % default: 5pt
	\begin{align*} 
		&\Sigma_{22}  \times 10^3 =\\
		&
		\left( \begin{array}{rrrrrrrrrr} 
			1500 & 124 & 814 & -104 & -179 & -223 & -731 & -189 & -400 & 242\\
			124 & 836 & 679 & 277 & 197 & -515 & -52.1 & -273 & 101 & 301\\
			814 & 679 & 1500 & 651 & 755 & -605 & -379 & -546 & -225 & 223\\
			-104 & 277 & 651 & 1960 & 720 & -782 & -299 & -775 & -180 & 506\\
			-179 & 197 & 755 & 720 & 2290 & -973 & 518 & -19.1 & -604 & -369\\
			-223 & -515 & -605 & -782 & -973 & 1290 & -400 & 412 & 314 & -420\\
			-731 & -52.1 & -379 & -299 & 518 & -400 & 1960 & 68.3 & 455 & -316\\
			-189 & -273 & -546 & -775 & -19.1 & 412 & 68.3 & 576 & -53.6 & -332\\
			-400 & 101 & -225 & -180 & -604 & 314 & 455 & -53.6 & 1030 & 265\\
			242 & 301 & 223 & 506 & -369 & -420 & -316 & -332 & 265 & 1090
		\end{array} \right).
	\end{align*}
\end{frame}

\end{document}